\documentclass[journal]{IEEEtran}
\usepackage{graphicx}
\usepackage{subfigure}  
\usepackage{multirow}			
\usepackage{booktabs}
\usepackage{times}  
\usepackage{helvet} 
\usepackage{courier}  
\usepackage[hyphens]{url}  
\usepackage{caption} 
\usepackage{amsmath,amsfonts,amsmath,amssymb}
\usepackage{bm}
\usepackage{CJKutf8}  
\usepackage{color}
\ifCLASSINFOpdf
\else
\fi
\begin{document}
	\begin{CJK*}{UTF8}{gbsn}
		%
		\title{Tri-Attention: Explicit Context-Aware Attention Mechanism for Natural Language Processing}
		%
		%
		%
		
		\author{Rui~Yu,~
			Yifeng~Li,~\IEEEmembership{Member,~IEEE,}
			Wenpeng~Lu,~\IEEEmembership{Member,~IEEE,}
			and~Longbing~Cao,~\IEEEmembership{Senior~Member,~IEEE}
			\thanks{The research work is partly supported by National Key R\&D Program of China under Grant No.2018YFC0831700, Australian Research Council Discovery under Grant No.DP190101079 and Future Fellowship under Grant No.FT190100734, Natural Science Foundation of Shandong under Grant No.ZR2022MF243 and Key Program of Science and Technology of Shandong Province under Grant No.2020CXGC010901. \textit{(Corresponding author: Wenpeng Lu.)}}
			\thanks{Rui Yu and Wenpeng Lu are with the Department
				of Computer Science and Technology, Qilu University of Technology (Shandong Academy of Sciences), Jinan 250353, China (e-mail: rui.yu1996@foxmail.com; wenpeng.lu@qlu.edu.cn).
				
				Yifeng Li 
				is with Department of Computer Science, Brock University, Niagara Region, ON
				L2S 3A1, Canada (e-mail: yli2@brocku.ca). 
				
				Longbing Cao is with the Data Science Institute, University of Technology Sydney, Sydney, NSW 2007, Australia (e-mail: longbing.cao@uts.edu.au). 
				
				
				}
			\thanks{Manuscript received October 15, 2022.}}
		
		%
		%

		\markboth{Journal of \LaTeX\ Class Files,~Vol.~14, No.~8, August~2015}%
		{Shell \MakeLowercase{\textit{et al.}}: Bare Demo of IEEEtran.cls for IEEE Journals}
		%



		\maketitle
		
		\begin{abstract}
		    In natural language processing (NLP), the context of a word or sentence plays an essential role. Contextual information such as the semantic representation of a passage or historical dialogue forms an essential part of a conversation and a precise understanding of the present phrase or sentence. However, the standard attention mechanisms typically generate weights using query and key but ignore context, forming a Bi-Attention framework, despite their great success in modeling sequence alignment. This Bi-Attention mechanism does not explicitly model the interactions between the contexts, queries and keys of target sequences, missing important contextual information and resulting in poor attention performance. Accordingly, a novel and general triple-attention (Tri-Attention) framework expands the standard Bi-Attention mechanism and explicitly interacts query, key, and context by incorporating context as the third dimension in calculating relevance scores. 
		    Four variants of Tri-Attention are generated by expanding the two-dimensional vector-based additive, dot-product, scaled dot-product, and bilinear operations in Bi-Attention to the tensor operations for Tri-Attention. Extensive experiments on three NLP tasks demonstrate that Tri-Attention outperforms about 30 state-of-the-art non-attention, standard Bi-Attention, contextual Bi-Attention approaches and pretrained neural language models\footnote{The source codes and data are available at https://github.com/yurui12138/Tri-Attention.}.
		\end{abstract}
		

		%
		\IEEEpeerreviewmaketitle

		\section{Introduction}
		\IEEEPARstart{A}{ttention} is the cognitive process of selectively concentrating on one task while ignoring others. It efficiently allocates the finite brain processing resources to more important tasks and distributes the resources to prioritized tasks \cite{Oberauer2019}. 
		In neural networks, an attention mechanism is an adaptive sparse method for encoding and modelling sequence interactions by the association scores between different elements \cite{hu2019introductory}.
		It has been widely applied in various tasks of natural language processing (NLP), including machine translation \cite{Alessandro2020Fixed}, text matching \cite{Lyu2021LET}, automatic question answering \cite{ye2020financial}, and reading comprehension \cite{Zhang2020DCMN}.
		
		\begin{figure}[t]
			\centering
			\includegraphics[width=0.9\columnwidth]{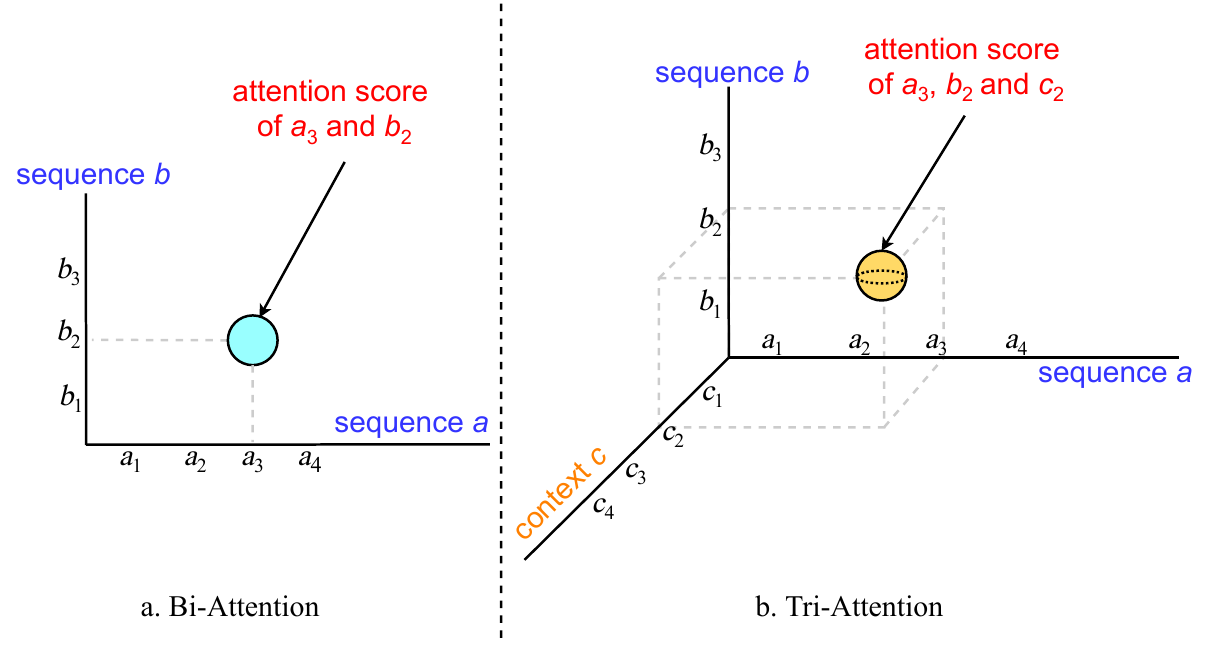}
			\caption{Comparison of the standard Bi-Attention mechanism with our proposed Tri-Attention mechanism. $a_i$ and $b_j$ represent two words from two sequences respectively, and $c_k$ denotes a contextual feature to them.}
			\label{toy_system}
		\end{figure}
		
		The attention mechanism learns an inner-representation of a sequence and the relationships between sequences, achieving great success in various tasks \cite{wang2020decoupled,basiri2021abcdm,Ding2020Self}. 
		However, it does not explicitly involve context as human cognition works, where context plays a critical role in human visual and language comprehension. Human brain has neural networks dedicated to reading and providing context from the environment in order to improve perceptual learning and visual interpretation \cite{Gilbert2000}. In natural language understanding, context is the key to derive and predict the meaning of a sentence \cite{Knoeferle2020}. It can improve language comprehensibility and how efficiently the comprehension is cognitively processed \cite{Willems2021}.
		Different from this human attention mechanism, the existing attention mechanisms in deep neural networks usually concentrate on word-level feature interactions but fail to fully account for the overall context of the word or sentence \cite{hu2021context}. 
		Accordingly, the attention matrix is calculated just by two individual tokens \textit{query} and \textit{key} extracted from sequences without a full consideration of their context, forming a standard query-key-based two-dimensional attention (we call it Bi-Attention for simplicity) framework to capture sequence interactions, as shown in Fig. \ref{toy_system}(a).
		This standard  Bi-Attention framework requires a deeper understanding of the data to obtain the inner representation of sequences and capture the intrinsic relationships between sequences \cite{storks2019recent}. Consequently, without involving context, the existing attention models may not effectively capture really important context-aware information \cite{hu2021context} and may produce  unsatisfied to inaccurate attention. 
		Some existing work has identified this design flaw in the  Bi-Attention mechanism and incorporated a contextual feature into the attention matrix. For example, the COIN and RE2 models align contextual features with attention to improve sentence representation learning, yielding better results than those without context \cite{hu2021context,Yang2019Simple}.

		Although the importance of context has been recognized in the existing work, context has only been employed as extra supporting information to sequence representation learning. No general frameworks are available to simulate human attention with context and explicitly capture the interactions between target and contextual sequences or information.  
		To manage this framework gap, we propose a triple-attention (Tri-Attention) framework to simulate human attention and explicitly capture interactions between sequences and between sequences and context. 
		Different from existing  Bi-Attention and contextual attention like COIN and RE2 models by integrating contextual features into queries and keys, our Tri-Attention mechanism directly involves context as the third dimension in quantifying sequence interactions. A query-key-context three-dimensional attention framework is formed for Tri-Attention, as shown in Fig. \ref{toy_system}(b).
		The standard Bi-Attention mechanism adopts \emph{query} and \emph{key} to calculate two-dimensional relevance  (e.g., dot product), forming an attention matrix. Then matrix operations are applied to the attention matrix and \emph{value} to obtain the attention embedding.
		Differently, the attention matrix in Tri-Attention captures three-dimensional interactions between \emph{query}, \emph{key}, and \emph{context}. To make \emph{value} consistent with the semantic space of the triple attention matrix, the relevance between \emph{context} and \emph{value}  is also calculated, producing \emph{contextual value}. 
		Finally, the attention embedding is obtained as a weighted linear combination of contextual value vectors.
		We test Tri-Attention-enabled networks for three independent NLP tasks, including retrieval-based English dialogue, Chinese sentence matching, and English reading comprehension. The results show that Tri-Attention outperforms most state-of-the-art approaches without attention, with Bi-Attention, contextual Bi-Attention, and pretrained language models with context. Furthermore, we evaluate the different performance of Bi-Attention, contextual Bi-Attention versus Tri-Attention and the effectiveness of Tri-Attention through a case study. 
		
		The main contributions of this work include:
			\begin{itemize}
				\item The proposed Tri-Attention mechanism expands the standard two-dimensional attention framework to explicitly involve and couple contextual information with query and key, hence the attention weights more sufficiently capture  context-aware sequence interactions. To the best of our knowledge, this is the first work on explicitly involving context (contextual features) and learning query-key-context interaction-based attention between sequences.  
				\item Tri-attention takes a general three-dimensional tensor framework, which can be instantiated into different implementations and applied to various tasks. We illustrate four variants by expanding the additive, dot-product, scaled dot-product and trilinear operations on query, key and context using tensor algebra for calculating Tri-Attention.
				\item Extensive experiments on three different NLP tasks and their real-world public datasets demonstrate the effectiveness of Tri-Attention. The Tri-Attention-enabled networks produce substantially better performance than about 30 state-of-the-art methods.
			\end{itemize}
		
		The rest of this paper is organized as follows. Section \ref{sec:relatedwork} introduces the related work. Section \ref{sec:background} discusses the background and preliminaries.  Section \ref{sec:method} proposes Tri-Attention mechanism and its variant implementations. Section \ref{sec:TAN Network} introduces the multi-task-shared Tri-Attention network. Section \ref{sec:experiments} demonstrates the performance of Tri-Attention by comparing it with state-of-the-art methods in terms of a variety of aspects. Lastly, Section \ref{sec:conclusion} concludes this work.
		
		\section{Related Work}
		\label{sec:relatedwork}
		In this section, we briefly review the related work on attention mechanisms and involving contextual information in attention, respectively.
		
		Human perception and visual processing may selective certain relevant parts of an image while ignoring other irrelevant information \cite{Oberauer2019}. 
		This human attention mechanism has inspired the neural attention models widely applied in computer vision \cite{Mnih2014Recurrent}.
		Attention has also shown success in neural NLP tasks, such as machine translation, automatic question answering, sentence matching, and word sense disambiguation, where attention usually captures the interactions between sequences. 
        For example, in machine translation, a recurrent attention   measures the weights of all heads in each Transformer layer to build more efficient neural machine translation models \cite{Zeng2021recurrent}. In \cite{Zhang21}, syntax-enhanced attention mechanisms SEAs improve syntactic-enhanced machine translation by involving a dependency mask bias and a relative local-phrasal position bias. In \cite{zhang2021self}, a source-target bilingual syntactic alignment SyntAligner aligns the syntactic structures of source and target sentences with border-sensitive span attention and then maximizes their mutual dependency for machine translation.       
        In automatic question answering, a Block-Skim attention mechanism measures the importance of tokens and blocks to skim unnecessary context, improving the Transformer performance \cite{guan2022block}. 
        In sentence semantic matching (SSM), A 3D CNN-based SSM model first constructs multi-dimensional representations for each sentence, then utilizes a 3D CNN and an attention mechanism to learn the interactive matching features \cite{Lu2021Sentence}. In word sense disambiguation (WSD), an extractive sense comprehension mechanism employs local self-attention to constraint attentions to be local, which allows to handle longer context without heavy computing burdens \cite{zhang2022word}.
  
		
		In NLP tasks, the existing Bi-Attention mechanisms calculate attention scores by focusing on local relevance matching
        at the token level without explicitly taking their overall context into account. Each element of the attention matrix is only computed based on two individual tokens \textit{query} and \textit{key}. This two-dimensional Bi-Attention mechanism ignores the interactions with and influence of their context, e.g., contextual words or sentences surrounding target words. On one hand, contextual information may involve important information and influence on the target, as demonstrated in other areas such as contextual and sequential recommender systems \cite{rashed2022context} and contextual matrix factorization \cite{zheng2022family}. On the other, the interactions and couplings between a target and its context form essential constituents of the target, as shown in attribute/feature interaction analysis \cite{zhang2021neural,cao2022beyond}, and multi-party interaction learning \cite{cao2022personalized,zhu2020unsupervised}. These prompt the potential and need for including context and target-context interactions into neural attention mechanisms. 

		In fact, various efforts aim to involve contextual information in  Bi-Attention mechanisms for NLP \cite{hu2021context,yang2019context,Ding2020Context}. 
		For example, Yang et al. \cite{yang2019context} directly incorporate contextual information into the representations of query and key with addition operations. 
		Ding et al. \cite{Ding2020Context} argue that the existing cross-attention was confused by the localness perception problem, which fails to adequately capture the whole context. A context-aware cross-attention method models both local and global contexts, which uses an interpolation gating mechanism to combine the original and local cross-attention.
		Hu et al. \cite{hu2021context} propose a context-aware attention network (COIN) for sentence matching. Its core component is a context-aware interaction block consisting of a context-aware cross-attention layer and a gate fusion layer, which consults contextual information to enable better alignments and blends the original and aligned representations with a gate connection. 
		
		In addition, contextual information or contextual attention has also been applied in various non-NLP tasks, such as image inpainting \cite{zeng2021cr,xiang2022deep}, video captioning \cite{song2022contextual}, 
        recommendation \cite{rashed2022context,pang2022heterogeneous}, and decision making \cite{Hallak2015,Benjamins2022}. 
		Regarding contextual attention for various vision and image-related tasks, for instance, in \cite{zeng2021cr}, contextual attention in CNN synthesizes both image structures and surrounding image features as references for image inpainting. 
        The selective contextual attention in \cite{liu2022deep} further learns pixel- and patch-level attentions from background regions and selectively utilizes contextual attention to enhance the original features. 
        The contextual attention network in \cite{song2022contextual} first extracts visual and textual features at each time step, then utilizes a contextual attention mechanism to capture more information for captioning. 
        In contextual attention-based recommendation, for example, attention-based influential contexts are modeled by involving heterogeneous relations for recommendation \cite{pang2022heterogeneous}. In \cite{rashed2022context}, the attention of context is measured and incorporated into sequential recommendation. These studies show the great potential of integrating context with attention for non-NLP tasks. 
        Furthermore, in contextual reinforcement learning (cRL), a contextual MDP (cMDP) extended the standard Markov Decision Process (MDP) to learn context-conditioned policies, which increases the robustness and generalization of RL models \cite{Hallak2015,Benjamins2022}.

        Although the above-mentioned work strives to implement contextual attention to model the influence from context, they do not directly capture target-context interactions, i.e., the interactions between \textit{context}, \textit{query} and \textit{key} in calculating attention scores. They still rely on the framework of standard query-key-based Bi-Attention mechanism, merely integrate contextual information and the original representation using addition operations or gating mechanisms, there are no-to-weak interactions between contextual information and the representations of \textit{query} and \textit{key}.

		To address the shortage of two-dimensional query-key attention  and the limitation of existing studies, we propose a query-key-context triple attention (Tri-Attention) framework. Tri-Attention uses tensor algebra techniques to explicitly involve contextual information and capture query-key-context interactions. In calculating the attention matrix, Tri-Attention incorporates contextual information and treats it equally with the Bi-Attention factors \emph{query}, \emph{key}, and \emph{value}. In this way, Tri-Attention expands Bi-Attention mechanisms by allowing contextual information to explicitly participate in calculating attention weights. 

        Consequently, the contextual information in Tri-Attention plays an essential role in capturing the interactions between sequences under their contexts. This also expands the contextual Bi-Attention mechanisms where context only plays a complementary role.

		\section{Background and Preliminaries}
        \label{sec:background}
		Here, we introduce tensor algebra and the design of the standard Bi-Attention mechanism. These lay the foundation of our proposed Tri-Attention.
		
		\subsection{Tensor Algebra}
		\label{sec:notation_tensor}
		Our Tri-Attention builds on tensor algebra to incorporate and interact context with query, key, and value and calculate attention weights on the query-key-context tensor. Therefore, here, we introduce the essential algebraic notations and concepts. We use a bold uppercase symbol to represent a matrix, e.g., $\bm{X}$; a bold lowercase for a vector (column), e.g., $\bm{x}$; and a lowercase or uppercase for a scalar, e.g., $x$ or $X$. A tensor is denoted by a bold calligraphic symbol, e.g., a three-dimensional real-valued tensor: $\bm{\mathcal{X}} \in \mathbb{R}^{I\times J \times K}$.
		
		The tensor or vector matrix multiplication \cite{Tensor_Kolda2009} is applied to calculate Tri-Attention weights. Given a tensor $\bm{\mathcal{X}}\in \mathbb{R}^{I_1 \times I_2 \times \cdots \times I_N}$, and a matrix $\bm{Y}\in \mathbb{R}^{J\times I_n}$, the $n$-mode product of $\bm{\mathcal{X}}$ and $\bm{Y}$ is defined as $\bm{\mathcal{X}} \times_n \bm{Y} = \bm{\mathcal{Z}} \in \mathbb{R}^{I_1\times I_2 \times \cdots \times I_{n-1} \times J \times I_{n+1} \times \cdots \times I_N}$, where
		\begin{align}
		\begin{small}
		\bm{\mathcal{Z}}_{i_1 i_2\cdots i_{n-1} j i_{n+1} \cdots i_N} =  \sum_{i_n=1}^{I_n} x_{i_1 i_2 \cdots i_{n-1} i_n i_{n+1} \cdots i_N} y_{j i_n}.
		\end{small}
		\label{eq:tensor_matrix}
		\end{align}
		Similarly, the $n$-mode product between $\bm{\mathcal{X}}$ and column vector $\bm{y}\in\mathbb{R}^{I_n}$ is written as $\bm{\mathcal{X}} \times_n \bm{y}^{\intercal} = \bm{\mathcal{Z}} \in \mathbb{R}^{I_1\times I_2 \times \cdots \times I_{n-1} \times 1 \times I_{n+1} \times \cdots \times I_N}$ with element-wise entry as: 
		\begin{align}
		\bm{\mathcal{Z}}_{i_1 i_2\cdots i_{n-1} 1 i_{n+1} \cdots i_N} =  \sum_{i_n=1}^{I_n} x_{i_1 i_2 \cdots i_n \cdots i_N} y_{i_n}.
		\label{eq:tensor_vector}
		\end{align}
		Oftentimes, the trivial $n$-th dimension in $\bm{\mathcal{Z}}$ is squeezed out such that $\bm{\mathcal{Z}} \in \mathbb{R}^{I_1\times I_2 \times \cdots \times I_{n-1} \times I_{n+1} \times \cdots \times I_N}$ is $(n-1)$-way. In this work, we keep this dimension for the convenience in formulating the Trilinear attention in Section \ref{sec:method}. 
		
		To formulate our idea, we use $\bm{C}$=$[\bm{c}_1,\bm{c}_2,\cdots,\bm{c}_J] \in \mathbb{R}^{D\times J}$ to represent the contextual matrix which contains $J$ context vectors. Each contextual vector is a column vector of length $D$, that is $\bm{c}_j \in \mathbb{R}^D$, $j=1,2,\cdots,J$. In this paper, a vector refers to a column vector by default. The key information is represented by matrix $\bm{K}$=$[\bm{k}_1,\bm{k}_2,\cdots,\bm{k}_I] \in \mathbb{R}^{D \times I}$ which includes $I$ key vectors of $D$ dimensions. Similarly, the corresponding value information is represented by $\bm{V}$=$[\bm{v}_1, \bm{v}_2, \cdots, \bm{v}_I] \in \mathbb{R}^{D\times I}$. In general, a query vector is denoted by $\bm{q} \in \mathbb{R}^D$. For $N$ query vectors, we use $\bm{Q}\in \mathbb{R}^{D\times N}$.

		\subsection{Standard Bi-Attention Mechanism}
		\label{subsec:Bi-Attention}
        The standard Bi-Attention mechanism customarily obtains interactive features by calculating the relevance score between sequences.
		It consists of three major steps.
		
		First, the relevance $F(\bm{q}, \bm{k}_i)$ between query $\bm{q}$ and key $\bm{k}_i$ is calculated:
		\begin{equation}
		\begin{aligned}
		F(\bm{q}, \bm{k}_i) = \mathrm{Similarity}(\bm{q}, \bm{k}_i), \quad i = 1,2, \cdots ,I,
		\end{aligned}
		\label{formula:atttention score}
		\end{equation}	

		Second, $F(\bm{q}, \bm{k}_i)$ is normalized by the softmax function:
		\begin{equation}
		{\alpha _i} = \frac{{{\exp{\big(F(\bm{q}, \bm{k}_i)\big)}}}}{{\sum\limits_{i' = 1}^I {\exp{\big(F(\bm{q}, \bm{k}_{i'})\big)}} }}, \qquad i = 1,2, \cdots ,I,
		\label{formula:normalized}
		\end{equation}	
		where ${\alpha _i}$ is the normalized attention weight.
		
		Lastly, the attention embedding is obtained by a weighted linear combination with value $\bm{v}_i$:
		\begin{equation}
		\begin{aligned}
		\bm{q}_{\text{new}} = \sum\limits_{i = 1}^I {{\alpha _i}{\bm{v}_i}} = \bm{V}  \bm{ \alpha} .
		\end{aligned}
		\end{equation}	
		
		In practice, the similarity measure is not unique, resulting in different attention mechanisms and designs. The four most commonly used similarity measure methods are additive (Add) similarity, dot-product (DP) similarity, scaled dot-product (SDP) similarity, and bilinear (Bili) similarity.
		
		Additive (Add) similarity \cite{Bahdanau2015Neural}:
		\begin{equation}
		\begin{aligned}
		F(\bm{q}, \bm{k}_i) = {\bm{p}^{\intercal}}\mathrm{tanh}(\bm{W}\bm{q} + \bm{U}\bm{k}_i), \quad i = 1,2, \cdots ,I,
		\end{aligned}
		\label{formula:concat score}
		\end{equation}
		
		Dot-product (DP) similarity \cite{Luong2015Effective}:
		\begin{equation}
		\begin{aligned}
		F(\bm{q}, \bm{k}_i) = {\bm{q}^{\intercal}}{\bm{k}_i}, \qquad i = 1,2, \cdots ,I,
		\end{aligned}
		\label{formula:dot score}
		\end{equation}
		
		Scaled dot-product (SDP) similarity \cite{Vaswani2017Attention}:
		\begin{equation}
		\begin{aligned}
		F(\bm{q}, \bm{k}_i) = \frac{{{\bm{q}^{\intercal}}{\bm{k}_i}}}{{\sqrt D }}, \qquad i = 1,2, \cdots ,I,
		\end{aligned}
		\label{formula:Scaled dot score}
		\end{equation}
		
		Bilinear (Bili) similarity \cite{Luong2015Effective}:
		\begin{equation}
		\begin{aligned}
		F(\bm{q}, \bm{k}_i) = {\bm{q}^{\intercal}}\bm{W}\bm{k}_i, \qquad i = 1,2, \cdots ,I,
		\end{aligned}
		\label{formula:general score}
		\end{equation}
		$\bm{W}$, $\bm{U}$ and $\bm{p}$ are learnable parameters.
		
		The main drawback of the standard Bi-Attention mechanisms lies in  that the relevance scores between sequences only build on the token representations of $\bm{q}$ and $\bm{k}_i$. No contextual information and the interaction between context and query and key are involved, which may miss important environmental information and cause inaccurate relevance.

		\section{Tri-Attention Mechanism}
		\label{sec:method}
		In this section, we introduce the proposed Tri-Attention mechanism. It extends the Bi-Attention mechanisms and builds on tensor operations for variant implementations of the query-key-context similarity.

		\subsection{Framework of Tri-Attention Mechanism}
		\label{subsec:Tri-Attention}
		
\begin{figure}[t]
			\centering
			\includegraphics[width=0.9\columnwidth]{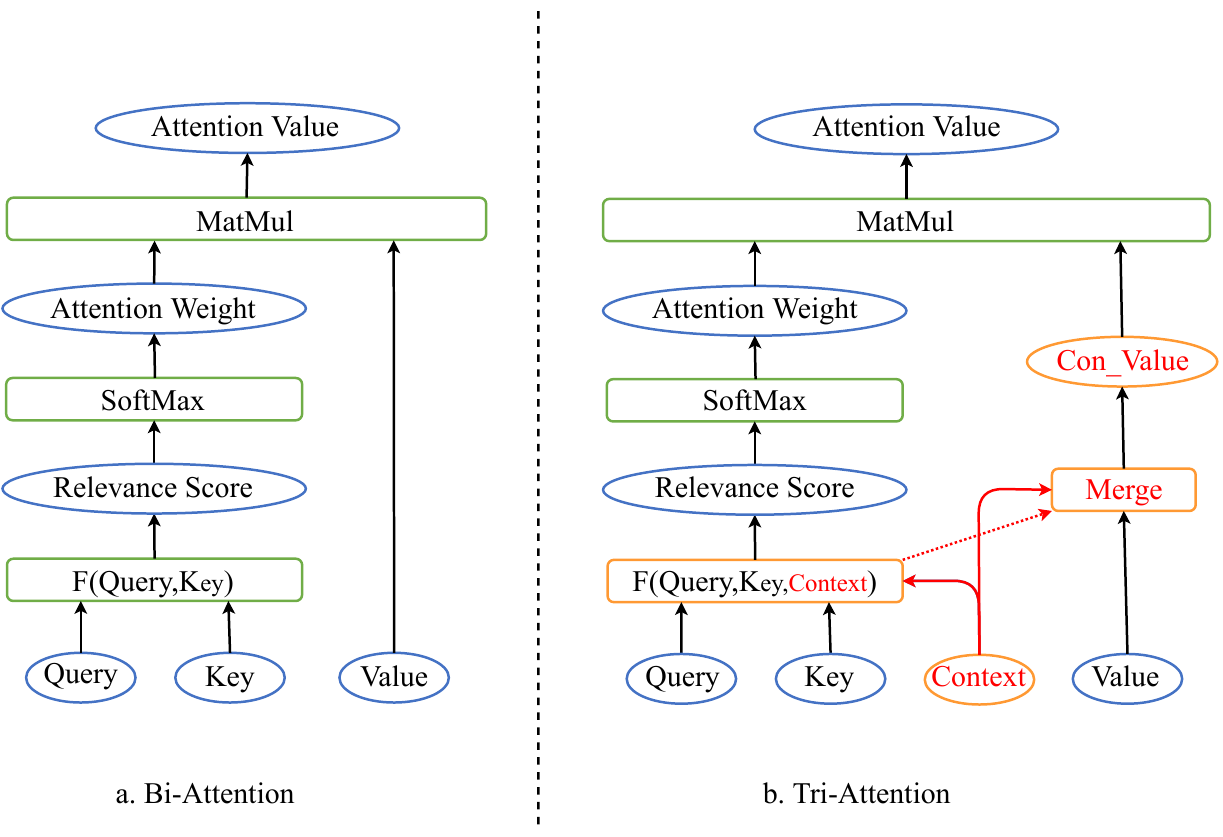}
			\caption{Comparison of the standard Bi-Attention versus Tri-Attention mechanisms. The elements in red refer to the new modules introduced by Tri-Attention, which adjust standard query-key-value to be context dependent. As a result, the calculation of contextual relevance scores is semantically consistent with the computation of contextual values.}
			\label{multil_atten}
		\end{figure}

		The Tri-Attention mechanism expands the standard query-key similarity-based Bi-Attention mechanism in Section \ref{subsec:Bi-Attention} to query-key-context tensor similarity, which thus explicitly involves contextual information and captures the interactions between target sequences and their contexts. Tri-Attention engages contextual features in calculating the relevance scores between sequences and then adjusts values to be context dependent. Bi-Attention can be viewed as a special case of Tri-Attention without the explicit dimension of context. Once \textit{context} is removed, Tri-Attention is degraded to Bi-Attention.
		Fig. \ref{multil_atten} illustrates the main ideas and processes of Bi-Attention versus Tri-Attention mechanism. The extension of Bi-Attention to Tri-Attention is marked in red for differentiating the two mechanisms. 
		
		In contrast to the standard three-step attention learning in Bi-Attention, Tri-Attention generalizes their first and third steps.
		In the first step, Tri-Attention first expands the similarity calculation  by involving contextual information to obtain context-dependent query-key relevance scores, forming \textit{contextual relevance score}. In the third step, Tri-Attention explicitly integrates context and value to produce context-dependent value, resulting in \textit{contextual value}. The resultant contextual value resides in the same semantic space as contextual relevance scores. Both capture extra contextual information and query-key-value-context interactions for a more informative but semantically consistent attention representation.
		
		\subsection{Contextual Relevance Score}
		To obtain relevance scores informative to the context of target sequences, we expand the relevance calculation methods including additive, dot-product, scaled dot-product, and bilinear similarity in Section \ref{subsec:Bi-Attention} by explicitly incorporating contextual information as a third dimension. This results in four contextual relevance similarity calculators: T-additive (TAdd), T-dot-product (TDP), T-scaled-dot-product (TSDP), and Trilinear (Trili) operations on the query-key-context tensor by applying the tensor-matrix or vector product operations defined in Eqs. \eqref{eq:tensor_matrix} and \eqref{eq:tensor_vector}. Here, prefix ``T-'' indicates its roles for Tri-Attention by involving the third dimension context into attention mechanisms. 
		
		\textit{T-additive (TAdd) similarity}:
		The T-additive similarity expands the additive similarity in Eq. \eqref{formula:concat score} by adding a context dimension $\bm{c}_j$ to the usual terms $\bm{q}$ and $\bm{k}_i$,  formulated below:
		\begin{equation}
		\begin{aligned}
		F(\bm{q}, \bm{k}_i, \bm{c}_j) = {\bm{p}^{\intercal}}\mathrm{tanh}(\bm{W}\bm{q} + \bm{U}\bm{k}_i + \bm{H}\bm{c}_j),\\
		i = 1,2,\cdots,I; \; j = 1,2,\cdots,J
		\end{aligned}
		\label{formula:concat score context}
		\end{equation}
		where $\bm{W}$, $\bm{U}$, $\bm{H}$, and $\bm{p}$ are learnable parameters.

		\textit{T-dot-product (TDP) similarity}:
		The T-dot-product similarity expands the standard query-key dot-product attention in Eq. \eqref{formula:dot score} for Bi-Attention to the query-value-context T-dot-product attention. T-dot-product attention replaces the inner product of query and value vectors by the \textit{contextual inner product} between the query, value, and context vectors, which is formulated below:
		\begin{equation}
		\begin{aligned}
		F(\bm{q},{\bm{k}_i},{\bm{c}_j}) &= \sum_{d=1}^D q_d k_{id} c_{jd}
		= \langle \bm{q}, \bm{k}_i, \bm{c}_j \rangle\\
		&= \bm{\mathcal{I}} \times_1 \bm{q}^{\intercal} \times_2 \bm{k}_i^{\intercal} \times_3 \bm{c}_j^{\intercal}\\
		&i = 1,2,\cdots,I; \; j = 1,2,\cdots,J
		\end{aligned},
		\label{formula:dot score context}
		\end{equation}
		where $\langle \bm{q}, \bm{k}_i, \bm{c}_j \rangle$ is the contextual inner product of three vectors; $\bm{q}^{\intercal}$, $\bm{k}_i^{\intercal}$, and $\bm{c}_j^{\intercal}$, which are treated as row vectors, i.e. of size $1\times D$.  
		$\times_1$, $\times_2$, and $\times_3$ are 1-mode, 2-mode, and 3-mode tensor-matrix or vector multiplication operators,  respectively. $\bm{\mathcal{I}}$ is a 3-way identity tensor of size $D \times D \times D$, where only the $(d,d,d)$-th element is 1 ($d=1,2,\cdots,D$), and others are 0. The resultant $F(\bm{q},{\bm{k}_i},{\bm{c}_j})$ is a scalar (i.e., of size $1\times 1 \times 1$).
		
		\textit{T-scaled-dot-product (TSDP) similarity}:
		Similarly, the scaled-dot-product attention in Eq. \eqref{formula:Scaled dot score} for Bi-Attention is generalized to T-scaled-dot-product. T-scaled-dot-product divides the contextual inner product by the squared root of the number of dimensions:
		\begin{equation}
		\begin{aligned}
		F(\bm{q},{\bm{k}_i},{\bm{c}_j}) &= \frac{\sum_{d=1}^D q_d k_{id} c_{jd}}{{\sqrt[] D }} = \frac{\langle \bm{q}, \bm{k}_i, \bm{c}_j \rangle}{{\sqrt[] D }} \\
		&= \frac{\bm{\mathcal{I}} \times_1 \bm{q}^{\intercal} \times_2 \bm{k}_i^{\intercal} \times_3 \bm{c}_j^{\intercal}}{{\sqrt[] D }}  \\ 
		&i = 1,2, \cdots ,I; \; j = 1,2, \cdots ,J
		\end{aligned}.
		\label{formula:Scaled dot score context}
		\end{equation}
		
		\textit{Trilinear (Trili) similarity}:
		With context involved, the bilinear form in  Eq. \eqref{formula:general score} is naturally extended to a multilinear, precisely trilinear, form using tensor-vector products, labeled as Trilinear: 
		\begin{equation}
		\begin{aligned}
		F(\bm{q},{\bm{k}_i},{\bm{c}_j}) &= \sum_{d=1}^D \sum_{d'=1}^D \sum_{d''=1}^D w_{dd'd''} q_d k_{id'} c_{jd''} \\ &= \bm{\mathcal{W}} \times_1 \bm{q}^{\intercal} \times_2 \bm{k}_i^{\intercal} \times_3 \bm{c}_j^{\intercal}\\
		&i = 1,2, \cdots ,I; \; j = 1,2, \cdots ,J
		\end{aligned},
		\label{formula:general score context}
		\end{equation}
		The learnable weight tensor  $\bm{\mathcal{W}}\in\mathbb{R}^{D\times D \times D}$ governs the interactions between any dimensions of the three vectors. T-dot-product and T-scaled-dot-product are special cases of this general form. A nice property of this contextual generalization is that the contextual relevance scores can be computed and stored in a tensor using query, key, and context matrices: $\bm{\mathcal{F}}(\bm{Q},\bm{K},\bm{C}) = \bm{\mathcal{W}} \times_1 \bm{Q}^{\intercal} \times_2 \bm{K}^{\intercal} \times_3 \bm{C}^{\intercal} \in \mathbb{R}^{N\times I\times J}$ where $N$, $I$, and $J$ are  the number of query vectors, key vectors, and context vectors, respectively. However, the size of $\bm{\mathcal{W}}$ grows in a cubic way. Hence, $\bm{\mathcal{F}}(\bm{Q},\bm{K},\bm{C})$ may not scale well. In practice, we can use an economic version as follows:
		\begin{equation}
		\begin{aligned}
		&F(\bm{q},{\bm{k}_i},{\bm{c}_j})
		= \langle \bm{W}\bm{q}, \bm{U}\bm{k}_i, \bm{H}\bm{c}_j \rangle\\
		&= \bm{\mathcal{I}} \times_1 (\bm{W}\bm{q})^{\intercal} \times_2 (\bm{U}\bm{k}_i)^{\intercal} \times_3 (\bm{H}\bm{c}_j)^{\intercal}\\ 
		&i = 1,2,\cdots,I; \; j = 1,2,\cdots,J
		\end{aligned},
		\label{formula:general score context2}
		\end{equation}
		where $\bm{W}$, $\bm{U}$, and $\bm{H}$ are learnable matrices. Concisely, for matrix inputs, we have $\bm{\mathcal{F}}(\bm{Q},\bm{K},\bm{C}) = \bm{\mathcal{I}} \times_1 (\bm{W}\bm{Q})^{\intercal} \times_2 (\bm{U}\bm{K})^{\intercal} \times_3 (\bm{H}\bm{C})^{\intercal} \in \mathbb{R}^{N\times I\times J}$.
		
		
	\subsection{Normalized Contextual Relevance Score}	
		Consistent with the Bi-Attention mechanism, the contextual relevance score $F(\bm{q},{\bm{k}_i},{\bm{c}_j})$ calculated by Eq. \eqref{formula:general score context2} for Tri-Attention is also normalized by the softmax function:
		\begin{equation}
		\begin{aligned}
		{\alpha _{ij}^c} &= \frac{{{\exp{\big(F(\bm{q},{\bm{k}_i},{\bm{c}_j})\big)}}}}{{\sum\limits_{i' = 1}^I \sum\limits_{j' = 1}^J \exp{\big(F(\bm{q},{\bm{k}_{i'}},{\bm{c}_{j'}})\big)} }} \\ & i = 1,2, \cdots ,I; \; j = 1,2, \cdots ,J
		\end{aligned}.
		\label{formula:normalized_new}
		\end{equation}
		Consequently, the attention weights for query $\bm{q}$ is represented by matrix $\bm{A}^c\in \mathbb{R}^{I\times J}$.
		
		\subsection{Contextual Value}
		The above contextual relevance scores contain contextual information and interactions between context, query, and key. To semantically match value to this contextual relevance score, we further integrate value with context using one of the following methods to obtain contextual value.
		
		
		
		\textit{Additive context-value integration:}
		\begin{equation}
		\begin{aligned}
		\bm{v}_{(i,j)}^c = \bm{v}_i + \bm{c}_j \quad
		i = 1,2, \cdots ,I; \ j = 1,2, \cdots ,J
		\end{aligned}.
		\label{formula:concat value context2}
		\end{equation}

		\textit{Multiplicative context-value integration:}
		\begin{equation}
		\begin{aligned}
		\bm{v}_{(i,j)}^c = \bm{v}_i * \bm{c}_j \quad
		i = 1,2, \cdots ,I; \ j = 1,2, \cdots ,J
		\end{aligned}
	\label{formula:dot value context}
		\end{equation}
		where $*$ is the Hadamard product operator. 
		
		
		\textit{Bilinear context-value integration:}
		\begin{equation}
		\begin{aligned}
		&\bm{v}_{(i,j)}^c = (\bm{U}'\bm{v}_i) * (\bm{H}' \bm{c}_j)\\
		&i = 1,2, \cdots ,I; \  j = 1,2, \cdots ,J.
		\end{aligned}
		\label{formula:general value context}
		\end{equation}
		
		In Eqs. \eqref{formula:concat value context2}-\eqref{formula:general value context}, $\bm{v}_{(i,j)}^c\in \mathbb{R}^D$. These methods generate the contextual value tensor $\bm{\mathcal{V}}^c\in \mathbb{R}^{I\times J \times D}$. 

		\begin{figure*}[htbp]
			\centering
			\includegraphics[width=0.95\textwidth]{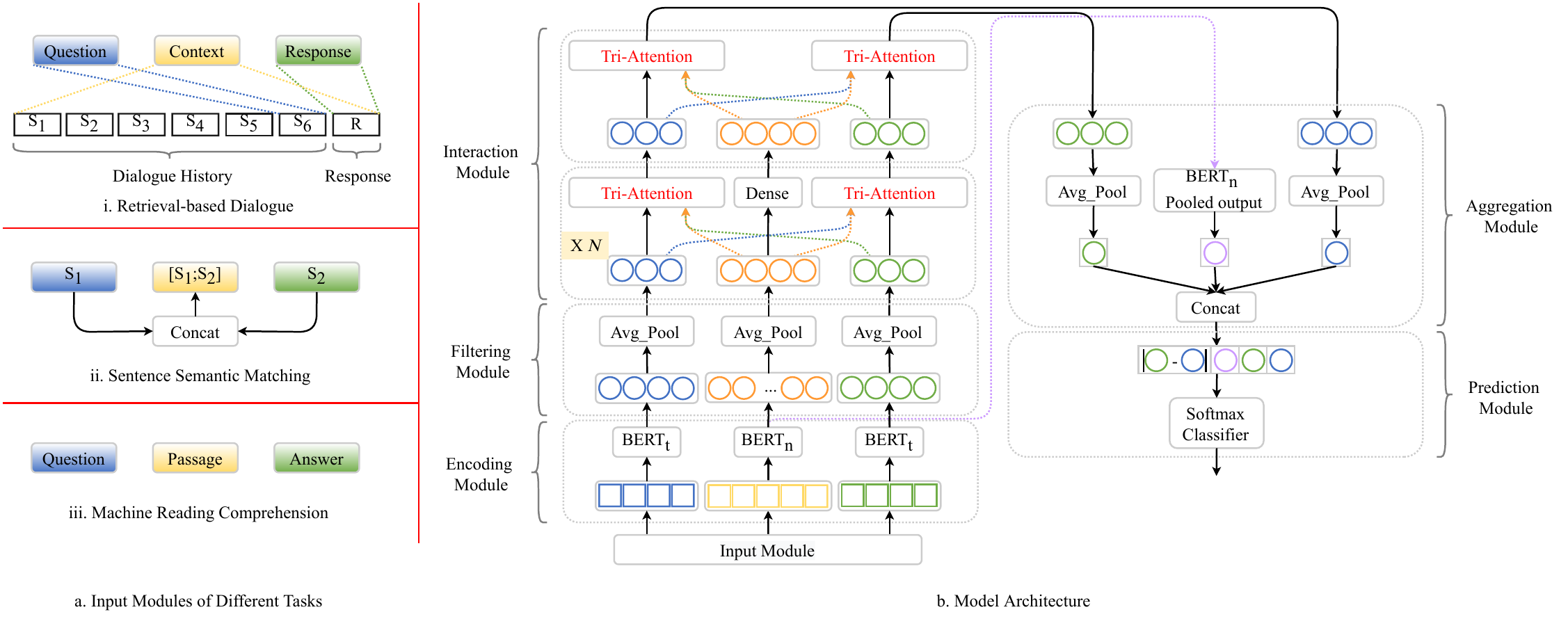}
			\caption{Shared architecture of Tri-Attention network (TAN) for three NLP tasks: left panel - input modules, right panel - model architecture.}
			\label{T-A}
		\end{figure*}

		\subsection{Contextual Attention Embedding}
		Since there are different approaches to obtain contextual attention weight matrix $\bm{A}^c$ by Eq. \eqref{formula:normalized_new} and value tensor $\bm{\mathcal{V}}^{c}$ per Eqs. \eqref{formula:concat value context2}-\eqref{formula:general value context}, for the semantic consistency, their operation may be consistent. For example, if T-additive in Eq. \eqref{formula:concat score context} is applied to obtain ${\alpha _{ij}^c}$, then additive context-value integration in Eq. \eqref{formula:concat value context2} should be correspondingly applied to obtain contextual value $\bm{v}_{(i,j)}^c$. Further, the new \textit{contextual attention embedding} $\bm{q}_{\mathrm{new}}^c$ corresponding to query $\bm{q}$ is:
		\begin{equation}
		\begin{aligned}
		\bm{q}_{\mathrm{new}}^c = \sum_{i=1}^I\sum_{j=1}^J \alpha_{ij}^c \bm{v}_{(i,j)}^c = \bm{V}^c \bm{\alpha}^c
		\end{aligned},
		\end{equation}
		where $\bm{q}_{\mathrm{new}}^c \in \mathbb{R}^D$,  $\bm{V}^c$ is mode-3 matricized from $\bm{\mathcal{V}}^c$: $\bm{V}^c$=$[\bm{v}^c_1, \bm{v}^c_2, \cdots, \bm{v}^c_{I*J}] \in \mathcal{R}^{D\times (I*J)}$, and $\bm{\alpha}^c$ is vectorized from $\bm{A}^c$. For $m$=$(i-1)I + j$, $\bm{v}^c_m$ in $\bm{V}^c$ corresponds to $\bm{v}^c_{(i,j)}$ in $\bm{\mathcal{V}}^c$ and $\alpha_m^c$ in $\bm{\alpha}^c$ is the same as $\alpha^c_{ij}$ in $\bm{A}^c$.

		\subsection{Choice of Contexts}
		Since contextual information varies over tasks and definitions, the acquisition of context is task-specific \cite{hu2021context, Yang2019Simple, Ding2020Context, Benjamins2022, Chen2020Neural}.
		As an example by following the practice in \cite{Chen2020Neural} for NLP tasks, we first choose the encoding results of BERT for different sequences as the preliminary contextual information. Then an average pooling operation is applied to obtain the final contextual features.
		
		More specifically, assume a task consists of several input sequences $S_1$, $S_2$, $\cdots$, and $S_K$.
		First, we segment each sequence according to the minimum granularity of the language underlying the inputs. For example, if the sequences are in Chinese, Chinese can be segmented in terms of Chinese characters. In contrast, if English is involved, the inputs should be segmented on the granularity of words.
		Then, we concatenate the segmented sequences with segments $\{SEP\}$ to a new sequence as follows:
		\begin{equation}
		\begin{aligned}
		S &= \{[CLS], S_1, [SEP], \cdots, [SEP], S_K, [SEP]\} \\
		&= \{[CLS], w_1^1, \cdots, w_1^{l_{S_1}}, [SEP], \cdots, [SEP], \\ & \qquad\qquad\qquad\qquad\qquad w_K^1, \cdots, w_K^{l_{S_K}}, [SEP]\},
		\end{aligned}
		\label{formula:sequences_concat}
		\end{equation}
		where $w_{k}^{l}$ is the $l$-th token in the $k$-th sequence.
		Finally, we feed this new sequence $S$ into BERT to obtain the contextual information, which consists of a list of vectors corresponding to all tokens respectively. It is worth noting that the output of BERT consists of two parts, and we use the first part as the contextual feature.

		\section{A Multi-task-shared Tri-Attention Network}
        \label{sec:TAN Network}

		Here, we apply Tri-Attention in diverse NLP tasks by constructing a multi-task-shared  Tri-Attention-enabled network (TAN). We test Tri-Attention for retrieval-based dialogue, sentence semantic matching, and machine reading comprehension. Accordingly, the shared architecture of our Tri-Attention network for these tasks is shown in Fig. \ref{T-A}. 
  
        TAN consists of two main panels. The left consists of three input modules corresponding to three NLP tasks, respectively. For each task, its input sequences and contexts specified according to the task-specific requirement, which are marked by different colors. The right panel implements the corresponding attention mechanism and learning tasks. It consists of five core modules: encoding module, filtering module, interaction module, aggregation module, and prediction module. 
        First, the encoding module employs BERT$_\text{t}$ to encode the two input sequences to obtain their feature representations, and utilizes BERT$_\text{n}$ to encode the context to obtain the contextual feature representation. Second, the filtering module trims the sequence length of the above representations to facilitate the following processing. Third, the interaction module applies Tri-Attention to engage three representations of input and context sequences and then couple them by the Tri-Attention mechanism. The stack number (i.e., $N$ in Fig.~\ref{T-A}) of interaction modules is adjustable per learning task requirements and based on the relevance score calculation methods, as shown in Table \ref{tabs:layers}. Furthermore, the feature aggregation module performs the average-pooling of the representations of two input sequences to filter them and then concatenates their representations with the pooling representation of the output of BERT$_\text{n}$, and the difference between representations of two input sequences. Lastly, a prediction module feeds the above four representations to a softmax function to predict the final response matching score.
        Below, in discussing the experimental settings for each task, we will further explain the task-specific settings to customize TAN for each task. 


		\section{Experiments}
        \label{sec:experiments}
		To verify the effectiveness of Tri-Attention mechanism in comparison with Bi-Attention, we conduct extensive experiments on three NLP tasks: retrieval-based dialogue, sentence semantic matching, and machine reading comprehension. Three public datasets are used to evaluate the results by comparing with different baseline methods for each task. In addition, we report the results of ablation study and case study.
		
		\subsection{Evaluation Tasks and Their Datasets}
		We evaluate Tri-Attention in three diverse NLP tasks: retrieval-based dialogue, sentence semantic matching, and machine reading comprehension. These tasks and their corresponding datasets are introduced below.
		\begin{itemize}
			\item \textit{Retrieval-based dialogue}. Given historical dialogue utterances, this task selects the correct response from multiple candidate responses. A commonly used data \textit{Ubuntu Corpus V1} \cite{Lowe2015Ubuntu} is tested here. It is a public dialogue dataset containing some 1 million multi-turn dialogues, with a total of over 7 million utterances and 100 million words. The ratio of positive versus negative instances in the training set is 1:1. The ratio of positive versus negative instances in the validation and test sets is 1:9. Table \ref{tabs:example_Dialogue} illustrates the samples from the Ubuntu Corpus V1 corpus. S$_1$-S$_6$ are historical dialogue utterances, and the candidate response utterances consist of a positive case and a negative case. 
			
			\begin{table}[htbp]\small
				\begin{center}	
					\begin{tabular}{l|c|l}
						\hline
						\hline
						& {\multirow{6}*{S$_1$}} & {hey guys i am trying to compile an}\\
						&  & {application \_\_path\_\_ went well and}\\
						&  & {it created a \_\_path\_\_ which takes}\\
						&  & {some arguments but i dont know}\\
						&  & {how to use this file ... can somebody}\\
						&  & {give me a hint}\\
						\cline{2-3}
						{Historical} & S$_2$ & {did n't \_\_path\_\_ created a makefile	}\\
						\cline{2-3}
						{dialogue} & S$_3$ & {ikonia take a look at this \_\_url\_\_	}\\
						\cline{2-3}
						{utterances} & S$_4$ & {it did but is does nothing	}\\
						\cline{2-3}
						& {\multirow{2}*{S$_5$}} & {it just mentions make *** no targets}\\
						&  & {specified and no makefile found stop ."}\\
						\cline{2-3}
						& {\multirow{2}*{S$_6$}} & {seems like \_\_path\_\_ uses this .. you}\\
						&  & {do n't have to run this manually}\\
						\hline
						& Positive & {look at the install file i think i just}\\
						Candidate & (label:1) & {have to pass the right arguments}\\
						\cline{2-3}
						{response} &  & {thanks a lot i think i have more to go}\\
						{utterances} & Negative & {on for my little project the syntax there }\\
						& (label:0) & {there is pretty easy to parse and all you}\\
						&  & {need is to make wget download that url}\\
						\hline\hline
					\end{tabular}
				\end{center}
				\caption{\label{tabs:example_Dialogue} Instances of retrieval-based dialogues \cite{Lowe2015Ubuntu}.} 
			\end{table}
			
			\item \textit{Sentence semantic matching}. This task decides whether two sentences share similar meaning. We use a large-scale Chinese corpus LCQMC \cite{liu2018lcqmc} for sentence semantic matching. It consists of 260,068 question pairs collected by Baidu Knows. There are three subsets: 238,766 question pairs for training, 8,802 question pairs for validation, and 12,500 question pairs for test. Table \ref{tabs:example_LCQMC} illustrates two of its instances in LCQMC.
			
			\begin{table}[htbp]\small
				\begin{center}
					\begin{tabular}{c|c|l}
						\hline\hline
						&    \multirow{2}*{S$_1$}& 哪首歌里有这句歌词 \\
						Positive&    &(En: Which song has this lyrics)	\\
						\cline{2-3}
						(label:1)&    \multirow{2}*{S$_2$}& 这句歌词是哪首歌的？\\
						&    &(En: Which song does this lyric come from?)	\\
						\hline
						&   \multirow{2}*{S$_3$}& 哪些浏览器可以看电影	\\
						Negative&    &(En: Which browser can play movies)\\
						\cline{2-3}
						(label:0)&    \multirow{2}*{S$_4$}& 什么浏览器可以下电影	\\  
						&    &(En: Which browser can download movies)  \\
						\hline\hline
					\end{tabular}
				\end{center}
				\caption{\label{tabs:example_LCQMC} Instances of sentence semantic matching data \cite{liu2018lcqmc}.}
			\end{table}
			
			
			\item \textit{Machine reading comprehension}. Given a passage and its corresponding question, this task identifies the correct answer from multiple candidate choices. RACE \cite{Lai2017RACE} is recognized as one of the largest and most difficult English datasets for multi-choice reading comprehension. It consists of two subsets: RACE-M and RACE-H, corresponding to the difficulty level for middle school and high school, respectively. Table \ref{tabs:example_RACE} illustrates this data. 
			
			\begin{table}[htbp]\small
				\begin{center}	
					\begin{tabular}{l|l}
						\hline
						\hline
						&  {Are you carrying too much on your back at}\\
						&  {school? You're not alone. Back experts in}\\
						&  {the USA are worried about that young students}\\
						&  {are having back and neck problems because}\\
						&  {they are carrying too much in their backpacks}\\
						{Passage} &  {...}\\
						&  {(3) The heaviest things should be packed}\\
						&  {closest to the back.}\\
						&  {(4) Bend  both knees when you pick up the}\\
						&  {pack, don't just bend over the waist}\\
						\hline
						{Question}& {The main idea of the passage is about \_\_.}\\
						\hline
						& {A. the problems made by rolling backpacks}\\
						Candidate & {B. the advantage of backpacks}\\
						choices & {C. the best backpacks for students}\\
						& {D. how to lighten students' backpacks}\\
						\hline
						Correct answer& {D}\\
						\hline\hline
					\end{tabular}
				\end{center}
				\caption{\label{tabs:example_RACE} Instances of machine reading comprehension data \cite{Lai2017RACE}} 
			\end{table}
			
		\end{itemize}
		
		\begin{table*}[h]\small
			\begin{center}	
				\begin{tabular}{l|l|c|c|c|c}
					\hline
					\hline
					Task & Dataset & {Tri-Attention$_{\text{TAdd}}$} & {Tri-Attention$_{\text{TDP}}$} & {Tri-Attention$_{\text{TSDP}}$}  &  {Tri-Attention$_{\text{Trili}}$}\\
					\hline
					Retrieval-based dialogue & Ubuntu Corpus V1 &3& 2 & 3 &4\\
					\hline
					Sentence semantic matching & LCQMC  &2& 3 & 3 &1 \\
					\hline
					Machine reading comprehension & RACE &4& 3 & 4 &1\\
					\hline\hline
				\end{tabular}
			\end{center}
			\caption{\label{tabs:layers} Number of Tri-Attention layers with different similarity operations in TAN for different NLP tasks and datasets.} 
		\end{table*}
		
		\subsection{TAN Settings}
		As shown in Fig. \ref{T-A}, the contextual information in Tri-Attention the concatenation of the outputs of BERT applied on input sequences, consistent with  \cite{Chen2020Neural}.
	    It is shown in \cite{Jawahar2019What} that the outputs of different layers of BERT can capture different sentence features. Inspired by this, our experiment shows that better performance can be achieved using the output of layer 1 as the representation of a single sequence.
		When the final representations of different sequences are obtained, we filter the representations by average pooling.
		Finally, inspired by \cite{Reimers2019Sentence}, we concatenate the representations of different sequences, the differential representations between sequences, and the pooled representation of BERT. The concatenate results are fed to softmax for classification.
		We also add the dropout strategy with dropout rate 0.1 following \cite{Nitish2014Dropout}.
		The above settings are applied to all three tasks: retrieval-based dialogue, sentence semantic matching, and machine reading comprehension.
		
		In addition, some hyperparameters are set for different tasks.
		The learning rate 1e-5 is used for retrieval-based dialogue, and 2e-5 for both sentence semantic matching and machine reading comprehension. The batch size for retrieval-based dialogue is 32, with 64 for sentence semantic matching and 16 for machine reading comprehension. 
		The cross-entropy loss is used for the objective function of all tasks. AdamW \cite{Ilya2019Decoupled} is used for the optimization of retrieval-based dialogue and sentence semantic matching, and BertAdam \cite{Jacob2019BERT} for machine reading comprehension. 
		In TAN, Tri-Attention is stackable, thus the number of its layers is adjusted for different tasks. Table \ref{tabs:layers} describes the number of layers for different tasks and datasets under different Tri-Attention similarity operations.


		\subsection{Task 1: Retrieval-based Dialogue}
		
		Here, we evaluate TAN with Tri-Attention mechanism against three categories of baselines for retrieval-based dialogue. The baselines consist of non-attention classic models, standard Bi-Attention-based neural networks with context or interaction, and pretrained neural language networks.
		
		\subsubsection{Baseline Methods}
		First, several non-attention classic methods are compared with TAN:
		\begin{itemize}
		    \item \textit{TF-IDF}: A statistical method obtains the relevance features of sentences per the frequency of characters \cite{Lowe2015Ubuntu}.
			\item \textit{RNN}: A simple RNN encodes utterances to obtain a feature representation and calculates the relevance score based on the representation \cite{Lowe2015Ubuntu}.
			\item \textit{CNN}: A CNN extracts utterance features and calculates the relevance representation of the features \cite{Kadlec2015Improved}.
			\item \textit{LSTM}: LSTM serves as the context encoder to extract utterance features and then calculates the relevance representation \cite{Kadlec2015Improved}.
		\end{itemize}
		
		Second, several advanced neural networks with standard Bi-Attention mechanisms with context or interaction are compared with TAN:
		\begin{itemize}
			\item \textit{SMN}: A sequential matching network selects responses to multi-turn conversations, which matches a response to each utterance in the context at multi-level granularities and utilizes RNN to accumulate the matching vectors to model the relations between utterances \cite{Wu2017Sequential}.
			\item \textit{DUA}: A self-matching attention routes the vital information in each utterance then matches a response to each refined utterance, followed by attention mechanism to aggregate matching vectors to obtain the final matching score \cite{Zhang2018Modeling}.
			\item \textit{DAM}: It constructs the representations of utterances at different granularities solely with stacked self-attention then calculates the matching score between the context and response by cross-attention \cite{Zhou2018Multi}.
			\item \textit{IoI}: An interaction-over-interaction network performs matching by stacking multiple interaction blocks to achieve deep interaction between responses and utterances \cite{Tao2019One}.
			\item \textit{ESIM}: A sequential matching model based only on chain sequence to select multi-turn responses, which involves an enhanced sequential inference model and soft-alignment attention \cite{Chen2019Sequential}.
			\item \textit{MSN}: A multi-hop selector network selects the relevant utterances as the context then calculates the matching score between the context and response \cite{Yuan2019Multi}. 
			\end{itemize}
		
		Lastly, we compare TAN with several pretrained neural language models:
		\begin{itemize}
			\item \textit{BERT}: A classical pretrained language model, which is fine-tuned for retrieval-based dialogue \cite{Jacob2019BERT}.
			\item \textit{RoBERTa-SS-DA}: A RoBERTa-based approach, where a speaker segmentation scheme and dialogue augmentation improve the performance \cite{Lu2020Improving}.
			\item \textit{BERT-DPT}: It enhances the BERT ability by post-training on domain-specific corpus to train contextualized representations that are missing in general corpus \cite{Whang2020Effective}.
			\item \textit{BERT-VFT}: A fine-tuned model based on BERT, which involves parameter-efficient transfer learning for various tasks \cite{Houlsby2019Parameter}.
			\item \textit{SA-BERT}: A speaker-aware BERT-based model, which perceives the change of speaker information and incorporates domain knowledge into pretrained BERT \cite{Gu2020Speaker}.
			\item \textit{UMS$_{\text{BERT+}}$}: Utterance manipulation strategies are used to select multi-turn responses, including utterance insertion, deletion, and search strategies, to address some deficiencies in existing BERT models \cite{Whang2021Response}.
			\item \textit{BERT-SL}: A context-response matching model based on pretrained language models, which introduces four self-supervised tasks to jointly train the response selection model in a multi-task manner \cite{Xu2021Learning}.
			\item \textit{BERT-UMS+FGC}: A fine-grained contrastive learning method for response selection, which generates better matching representation at finer granularity to select positive responses \cite{Li2021Small}.
		\end{itemize} 

		\subsubsection{TAN Settings}
		The TAN with Tri-Attention in Fig. \ref{T-A} is specified as follows for retrieval-based dialogue. 
		The inputs shown on the left panel consist of question, response, and context. A question is the last sentence in the dialogue history and a response is the original response in the dataset. The context is the concatenation of dialogue and response \cite{Chen2020Neural}. 
        The model on the right panel has five core modules, as described in Section \ref{sec:TAN Network}. 
        The performance is evaluated by  R$_{n}@k$, which denotes whether the top-$k$ retrieved responses from the $n$ candidate responses contain the right responses.  

		\subsubsection{Performance Evaluation}
		The performance of TAN against all baseline models is shown in Table \ref{tabs:result_ubuntu}. We obtain the following observations. 
			
		First, compared with the non-attention classic methods, the advantage of TAN is very significant. The classic methods model sentences only using observable word features and simple interaction patterns to evaluate each candidate response. They are unable to effectively utilize latent semantic information and learn complex interaction patterns. Thanks to the powerful pretrained language model and the Tri-Attention mechanism, TAN learns more sophisticated semantic features than these baselines. The learned contextual features are beneficial for representing utterances and responses, thus measuring their relevance more accurately.
		
		Second, compared with Bi-Attention and interaction-based networks, TAN also performs much better. This is probably because TAN employs the pretrained BERT$_{\text{base}}$ model trained on large corpus, which enhances our model with great prior knowledge. In contrast, the benchmarks do not involve the vast amount of prior knowledge.
		
		Third, TAN also outperforms the pretrained BERT, which is the state-of-the-art language model with self-attention mechanism. Although BERT gains rich prior knowledge pretrained on a vast amount of textual data, it may not capture interactions and contexts between utterances and responses for this task. In addition, BERT  pretrained on a general corpus may not be adaptive to this domain-specific dialogue dataset. In contrast, TAN applies Tri-Attention to explicitly capture contexts and adopts the post-training strategy in \cite{Whang2020Effective} on the task dataset to optimize its specificity and adaptability.
			
		Lastly, TAN with Tri-Attention consistently outperforms all pretrained BERT variants. Although most baselines also perform post-training, they may fail to incorporate contextual information and accurately capture interactive features between utterances and responses. In contrast, Tri-Attention fully considers the context in its attention mechanism to determine the relevance between utterances and responses, contributing to the improved performance of TAN over the BERT variants.

		\begin{table}[t]\small
			\begin{center}	
				\begin{tabular}{l|ccc}
					\hline
					\hline
					Methods & {R$_{10}$@1} & {R$_{10}$@2} & {R$_{10}$@5} \\
					\hline
					TF-IDF &41.0 &54.5 &70.8  \\
					RNN  & 40.3 &54.7 &81.9  \\
					CNN &54.9& 68.4 &89.6  \\
					LSTM &63.8 &78.4 &94.9  \\
					\hline
					SMN  &72.6 &84.7 &96.1  \\
					DUA & 75.2 &86.8 &96.2  \\
					DAM &76.7 & 87.4 & 96.9  \\
					IoI &79.6 & 89.4 & 97.4  \\
					ESIM &79.6 & 89.4 & 97.5  \\
					MSN &80.0 & 89.9 & 97.8  \\
					\hline
					BERT  &80.8 & 89.7 & 97.5  \\
					RoBERTa-SS-DA & 82.6 & 90.9 & 97.8  \\
					BERT-DPT &85.1 & 92.4 & 98.4  \\
					BERT-VFT &85.5 & 92.8 & 98.5  \\
					SA-BERT &85.5 & 92.8 & 98.3  \\
					UMS$_{\text{BERT+}}$ &87.5 & 94.2 & 98.8  \\
					BERT-SL &88.4 & 94.6 & 99.0  \\
					BERT-UMS+FGC &\textbf{88.6} & \textbf{94.8} & \textbf{99.0}  \\
					\hline
					\multirow{2}*{TAN$_{\text{TAdd}}$}  &90.5 & 95.8 & 99.2  \\	
					&($\pm 0.3$)  & ($\pm 0.06$) &($\pm 0.05$)  \\
					\multirow{2}*{TAN$_{\text{TDP}}$}  &90.3 & \textbf{95.9} & \textbf{99.3}  \\	
					&($\pm 0.1$)  & ($\pm 0.3$) &($\pm 0.06$)  \\
					\multirow{2}*{TAN$_{\text{TSDP}}$}  &\textbf{90.6} & 95.7 & 99.2  \\
					&($\pm 0.3$)  & ($\pm 0.05$) &($\pm 0.04$)  \\
					\multirow{2}*{TAN$_{\text{Trili}}$} &90.1 & 95.7 & \textbf{99.3}  \\
					&($\pm 0.08$)  & ($\pm 0.09$) &($\pm 0.04$)  \\
					\hline\hline
				\end{tabular}
			\end{center}
			\caption{\label{tabs:result_ubuntu} Retrieval-based dialogue: Experimental results on the Ubuntu Corpus V1 corpus. The values associated with $\pm$ in brackets are standard deviation. Mean and standard deviation results are averaged over five runs of each model. Four TAN variants of different tensor operation-based Tri-Attention mechanisms: TAdd - T-Additive, TDP - T-Dot-Product, TSDP - T-Scaled-Dot-Product, and Trili - Trilinear.}
		\end{table}		
		
		\subsection{Task 2: Sentence Semantic Matching}
		We further test TAN with Tri-Attention mechanism against three categories of baselines for sentence semantic matching. The baselines consist of non-attention classic networks, more advanced networks with relations or standard Bi-Attention-based context, and pretrained neural language networks.
		
		\subsubsection{Baseline Methods}
		First, several non-attention classic networks are compared to TAN with Tri-Attention: 
		\begin{itemize}
			\item \textit{WMD$_{\text{char}}$} and \textit{ WMD$_{\text{word}}$}: They apply the Wasserstein distance to the matching degree between two sentences in terms of character and word, respectively \cite{liu2018lcqmc}.
			\item \textit{CNN$_{\text{char}}$} and \textit{ CNN$_{\text{word}}$}: They utilize a CNN to encode two sentences for their corresponding sentence representations, which are then concatenated to predict the matching degree with a softmax classifier \cite{liu2018lcqmc}.
			\item \textit{BiLSTM$_{\text{char}}$} and \textit{BiLSTM$_{\text{word}}$}: They are similar with the former CNN method replacing CNN by a BiLSTM component \cite{liu2018lcqmc}.
		\end{itemize}
			
		Second, more advanced networks with relations or standard Bi-Attention-based context are compared with TAN embedded with Tri-Attention mechanism: 
		\begin{itemize}
			\item \textit{BiMPM$_{\text{char}}$} and \textit{BiMPM$_{\text{word}}$}: They employ bilateral multi-perspective matching to determine the semantic consistency between sentences; BiLSTM learns sentence representation, matches two sentences from multi-perspectives, aggregates the matching results, and makes prediction by a dense layer \cite{Wang2017Bilateral}.
			\item \textit{MSEM}: A connected graph describes the relations between sentences and realizes a neural architecture for multi-task learning including both sentence matching and classification \cite{Huang2019Multi}.
			\item \textit{GMN}: A neural graph matching network is fed with all possible segmentation paths to form word lattice graphs and learns graph-based representations of sentences \cite{Chen2020Neural}.
			\item \textit{COIN}: A cross-attention mechanism combines contextual information aligning sequences, with aligned representations interpolated by a gate fusion layer \cite{hu2021context}.
			\item \textit{3DSSM}: A 3D CNN captures  temporal  and multi-granular features of sentences and generates matching representation  \cite{Lu2021Sentence}.
		\end{itemize}
		
		Lastly, we compare TAN to several pretrained neural language networks:
		\begin{itemize}
			\item \textit{BERT-Chinese}: A Chinese BERT model \cite{Jacob2019BERT}.
			\item \textit{BERT-wwm}: A Chinese BERT with an entire word masking mechanism  during pretraining \cite{Pre2019Cui}.
			\item \textit{BERT-wwm-ext}: A variant of BERT-wwm with more training data and steps \cite{Pre2019Cui}.
			\item \textit{ERNIE}: It learns language representation enhanced by knowledge masking strategies and entity- and phrase-level masking \cite{sun2019ernie}.
			\item \textit{K-BERT}: A model enhances BERT with HowNet by introducing soft position and visible matrix during fine-tuning and inference phases \cite{Liu2020BERT}.
			\item \textit{GMN-BERT}: A neural graph matching network with multi-granular input information \cite{Chen2020Neural}.
		\end{itemize}
		
		\subsubsection{TAN Settings}
		The TAN with Tri-Attention mechanism is customized for sentence semantic matching. Its architecture is composed of four core modules as shown in Fig.~\ref{T-A}. Compared to the TAN variant for retrieval-based dialogue, the filtering module is not necessary. This is because the filtering module trims too long sequences, while the sequence length in this task is acceptable. Since the input of sentence semantic matching consists of only two sentences, the contextual feature representation is learned by feeding connected results of question and answer into BERT$_\text{n}$ \cite{Chen2020Neural}. 

        \subsubsection{Performance Evaluation}
		The results of TAN for sentence semantic matching compared with all baseline models are shown in Table \ref{tabs:result_lcqmc}. We obtain the following observations.
		
		First, TAN significantly outperforms all non-attention classic methods, similar to that for retrieval-based dialogue. The classic networks are unable to effectively capture latent semantic information and complex interaction patterns. In contrast, our Tri-Attention-based TAN  combines the advantages of pretrained language model with contextual attention.
		
		Second, TAN also beats advanced networks with interaction or Bi-Attention mechanism. Similar to retrieval-based dialogue, TAN gains advantages from the prior knowledge learned by the pretrained model on large-scale corpus and contextual attention captured by Tri-Attention.

		Further, TAN performs better than pretrained BERT variants. The BERT variants including BERT-wwm, BERT-wwm-ext and ERNIE were trained on large data with more subtle training techniques. In contrast, TAN outperforms these baselines without the involvement of large data or specific training techniques. This means that TAN not only outperforms these powerful BERT variants in terms of accuracy and F$_1$-score but also involves less training costs. 
		Additionally, both K-BERT and GMN-BERT are the latest BERT variants for sentence semantic matching. Their performance is still inferior to the best results of TAN$_{\text{TAdd}}$ and TAN$_{\text{TSDP}}$.
		
		Lastly, TAN$_{\text{TAdd}}$, TAN$_{\text{TDP}}$, and TAN$_{\text{TSDP}}$ outperform all baselines in terms of accuracy, and TAN$_{\text{TAdd}}$ and TAN$_{\text{TSDP}}$ achieve better performance than baselines 
		in terms of F$_1$-score. These results verify the effectiveness of our model. TAN builds on BERT by replacing its self-attention mechanism with Tri-Attention mechanism, whose four similarity implementations make 1-2\% improvement in terms of accuracy compared to the original BERT. 
		This indicates the Tri-Attention mechanism plays a core role in making TAN better than BERT, i.e., beating the standard Bi-Attention mechanism.

		\begin{table}[t]\small
			\begin{center}	
				\begin{tabular}{l|cc}
					\hline
					\hline
					{Methods} & {Accuracy} & {F$_1$-score}\\
					\hline
					WMD$_{\text{char}}$  &70.6 &73.4 \\
					WMD$_{\text{word}}$ & 60.0 &70.8\\
					CNN$_{\text{char}}$  &71.8 &75.2 \\
					CNN$_{\text{word}}$ & 72.8 &75.7\\
					BiLSTM$_{\text{char}}$ &73.5& 77.5\\
					BiLSTM$_{\text{word}}$ &76.1 &78.9  \\
					\hline
					BiMPM$_{\text{char}}$ &83.4&85.0\\
					BiMPM$_{\text{word}}$ &83.3 &84.9  \\
					MSEM &85.7& -\\
					GMN &84.6 &86.0  \\
					COIN &85.6 &86.5 \\
					3DSSM &85.7 &86.4  \\
					\hline
					BERT  &85.73 &86.86 \\
					BERT-wwm & 86.80 &87.78\\
					BERT-wwm-ext &86.68& 87.71\\
					ERNIE &87.04 &\textbf{88.06} \\
					K-BERT &87.10 &-  \\
					GMN-BERT &\textbf{87.30} &88.0  \\
					\hline
					TAN$_{\text{TAdd}}$  &\textbf{87.49} ($\pm 0.47$)  & \textbf{87.95} ($\pm 0.27$)\\
					TAN$_{\text{TDP}}$  &87.25 ($\pm 0.11$)  &87.83 ($\pm 0.06$)\\
					TAN$_{\text{TSDP}}$  &87.23 ($\pm 0.58$)  &87.80 ($\pm 0.29$)\\
					TAN$_{\text{Trili}}$  &86.72 ($\pm 0.05$)  &87.38 ($\pm 0.02$)\\
					\hline\hline
				\end{tabular}
			\end{center}
			\caption{\label{tabs:result_lcqmc}Sentence semantic matching: Experimental results on the LCQMC data. Mean and standard deviation results are averaged over five runs of each model.} 
		\end{table}

		\subsection{Task 3: Machine Reading Comprehension}
		The last NLP task for evaluating TAN with Tri-Attention against the baselines is machine reading comprehension. We evaluate this in terms of three sets of baselines: single models, ensemble models, and pretrained lanaguage neural networks.
		
        \subsubsection{Baseline Methods}
		First, we compare TAN with single model-based methods:
		\begin{itemize}
			\item \textit{Stanford AR}: The standard Bi-Attention mechanism obtains the question-related passage representation, which is then coupled with a candidate option to obtain the relevance score \cite{Lai2017RACE}.
			\item \textit{GA Reader}: A gated attention mechanism captures the passage information related to a problem, which is coupled with candidate options \cite{Lai2017RACE}.
			\item \textit{ElimiNet}: An elimination gate discards irrelevant options, refines passage representations, and selects the most suitable option by a selection module \cite{Parikh2018ElimiNet}.
			\item \textit{HAF}: A hierarchical attention mechanism captures the interactions between passages, questions and candidate options \cite{Zhu20118Hierarchical}.
			\item \textit{MUSIC}: It dynamically applies different matching strategies to different questions and applies a multi-hop reasoning method to reach the right answer \cite{Xu2017Towards}.
			\item \textit{Hier-Co-Matching}: A co-matching strategy matches a passage to a question and a passage to a candidate answer, and then leverages a hierarchical LSTM to encode co-matching states for the relevance representation \cite{Wang2018Matching}.
		\end{itemize}
		
		Second, we	compare TAN with ensemble models:
		\begin{itemize}
			\item \textit{GA Reader (6-ensemble)}: An ensemble method based on multi-hop gated attention mechanism between passages and questions \cite{Lai2017RACE}.
			\item \textit{ElimiNet (6-ensemble)}: An ensemble method with a multi-hop mechanism eliminates candidate answers \cite{Parikh2018ElimiNet}.
			\item \textit{GA + ElimiNet (12-ensemble)}: It integrates GA Reader (6-ensemble) and ElimiNet (6-ensemble) \cite{Chen2019Convolutional}.
			\item \textit{Dynamic Fusion Network (9-ensemble)}: An ensemble model based on multi-strategy inference for a comprehension architecture \cite{Xu2017Towards}.
			\item \textit{CSA Model + ELMo (9-ensemble)}: It integrates spatial convolution attention mechanism with pretrained language model ELMo \cite{Chen2019Convolutional}.
		\end{itemize}
		
		Lastly, we compare TAN with pretrained  language models:
		\begin{itemize}
			\item \textit{BERT$_{\text{base}}$}: It is the most commonly used pretrained language model structure \cite{Jacob2019BERT}.
			\item \textit{BERT$_{\text{base}}$ + DCMN}: A reading comprehension model based on BERT$_{\text{base}}$ and  a dual co-matching network mechanism \cite{Zhang2020DCMN}.
		\end{itemize}

        \subsubsection{TAN Settings}
		We here customize TAN with Tri-Attention for machine reading comprehension. As shown in Fig. \ref{T-A}, its model architecture is similar to that of retrieval-based dialogue. However, this task involves three input sequences. Similar to sentence semantic matching, contextual feature representation is obtained by feeding the concatenation of question, answer, and passage representations into BERT$_\text{n}$.


		\subsubsection{Performance Evaluation}
		The results of TAN with Tri-Attention for machine reading comprehension in comparison with all baselines are reported in Table \ref{tabs:result_race}. Again, TAN with Tri-Attention outperforms most baselines, verifying their effectiveness.
		
		First, TAN significantly outperforms all single-model-based methods. This is probably because these single models have a single structure which cannot model the semantic information within a text and the semantic relations between texts. TAN embedded with a pretrained language model and contextual attention shows their empower.
		
		Second, our approach is clearly superior to all ensemble ones. The first four ensemble models: GA Reader (6-ensemble), ElimiNet (6-ensemble), GA + ElimiNet (12-ensemble), and Dynamic Fusion Network (9-ensemble), do not apply pre-training, which may be the main reason for their low performance. The last ensemble model CSA Model + ELMo (9-ensemble) utilizes the pretrained language model ELMo \cite{Peters2018Deep} to enhance its performance. Although this model has been greatly improved compared with the previous four models, its performance is still much lower than TAN. This may be due to two reasons: the advantage of pretrained BERT in our model and the contextual enhancement by Tri-Attention.

		Lastly, similar to the above conclusions in other tasks, TAN with pretrained BERT beats the original BERT with self-attention by explicitly capturing contextual interactions with queries and keys. Specifically, the TAN with additive Tri-Attention similarity slightly outperform BERT$_{\text{base}}$+DCMN, which is specific for machine reading comprehension and relies on strategies and skills used by humans to complete reading comprehension tasks. In contrast, as a general attention mechanism, TAN and Tri-Attention are applicable to machine reading comprehension and other NLP tasks.
		
		\begin{table}[t]\small
			\begin{center}	
				\begin{tabular}{l|c}
					\hline
					\hline
					{Methods} & {Accuracy} \\
					\hline
					Stanford AR &43.3   \\
					GA Reader &44.1   \\
					ElimiNet &44.7   \\
					HAF &47.2  \\
					MUSIC &47.4   \\
					Hier-Co-Matching &50.4   \\
					\hline
					GA Reader (6-ensemble)  &45.9  \\
					ElimiNet (6-ensemble) & 46.5 \\
					GA + ElimiNet (12-ensemble) &47.2   \\
					Dynamic Fusion Network (9-ensemble) &51.2   \\
					CSA Model + ELMo (9-ensemble) &55.0   \\
					\hline
					BERT$_{\text{base}}$ &\textbf{65.0}   \\
					BERT$_{\text{base}}$+DCMN &\textbf{67.0}   \\
					\hline
					TAN$_{\text{TAdd}}$  & \textbf{67.5} ($\pm 0.04$) \\
					TAN$_{\text{TDP}}$  & 66.9 ($\pm 0.13$) \\
					TAN$_{\text{TSDP}}$  &66.7 ($\pm 0.07$)  \\
					TAN$_{\text{Trili}}$  & 66.1 ($\pm 0.14$) \\
					\hline\hline
				\end{tabular}
			\end{center}
			\caption{\label{tabs:result_race}Machine reading comprehension: Experimental results on RACE. Means and standard deviations are averaged over five runs of each model.} 
		\end{table}

		\begin{table*}[t]\small
			\begin{center}
				\begin{tabular}{l|ll|lll}
					\hline\hline
					\multirow{2}*{Methods} & \multicolumn{2}{c|}{LCQMC} & \multicolumn{3}{c}{Ubuntu Corpus V1 Corpus} \\
					\cline{2-6}
					& Accuracy & F$_1$-score & {R$_{10}$@1} & {R$_{10}$@2} & {R$_{10}$@5} \\
					\hline
					Tri-Attention$_\text{TAdd}$	&\textbf{87.49} ($\pm 0.47$)  & \textbf{87.95} ($\pm 0.27$) &\textbf{90.5} ($\pm 0.3$) & \textbf{95.8} ($\pm 0.06$) & \textbf{99.2} ($\pm 0.05$)	\\
					Bi-Attention$_\text{Add}$	& 86.77 ($\pm 0.28$)  & 87.37 ($\pm 0.16$) & 89.9 ($\pm 0.4$) & 95.5 ($\pm 0.2$) & \textbf{99.2} ($\pm 0.09$)	\\
					\hline
					Tri-Attention$_\text{TDP}$	&\textbf{87.25} ($\pm 0.11$)  &\textbf{87.83} ($\pm 0.06$) &\textbf{90.3} ($\pm 0.1$) & \textbf{95.9} ($\pm 0.3$) & \textbf{99.3} ($\pm 0.06$)	\\
					Bi-Attention$_\text{DP}$	& 86.52 ($\pm 0.09$)  & 87.20 ($\pm 0.06$)& 89.8 ($\pm 0.2$) & 95.5 ($\pm 0.1$) & 99.2 ($\pm 0.04$)	\\
					\hline
					Tri-Attention$_\text{TSDP}$	&\textbf{87.23} ($\pm 0.58$)  &\textbf{87.80} ($\pm 0.29$) &\textbf{90.6} ($\pm 0.3$) & \textbf{95.7} ($\pm 0.05$) & \textbf{99.2} ($\pm 0.04$)	\\
					Bi-Attention$_\text{SDP}$	& 86.61 ($\pm 0.09$)  & 87.29 ($\pm 0.04$) & 89.8 ($\pm 0.3$) & 95.5 ($\pm 0.2$) & 99.1 ($\pm 0.05$)	\\
					\hline
					Tri-Attention$_\text{Trili}$	&\textbf{86.72} ($\pm 0.05$)  &\textbf{87.38} ($\pm 0.02$) &\textbf{90.1} ($\pm 0.08$) & \textbf{95.7} ($\pm 0.09$) & \textbf{99.3} ($\pm 0.04$)	\\
					Bi-Attention$_\text{Bili}$	& 85.84 ($\pm 0.11$)  & 86.72 ($\pm 0.06$) & 89.7 ($\pm 0.09$)  & 95.5 ($\pm 0.1$) & 99.2 ($\pm 0.03$)	\\
					\hline\hline
				\end{tabular}
				\caption{\label{tabs:Effectiveness Ablation} Results of effectiveness of Bi-Attention vs Tri-Attention mechanisms on LCQMC and Ubuntu Corpus V1 Corpus. Means and standard deviations are averaged  over five runs.}
			\end{center}
		\end{table*}
		
		\begin{table*}[h]\small
			\begin{center}
				\begin{tabular}{l|ll|lll}
					\hline\hline
					\multirow{2}*{Methods} & \multicolumn{2}{c|}{LCQMC} & \multicolumn{3}{c}{Ubuntu Corpus V1 Corpus} \\
					\cline{2-6}
					& Accuracy & F$_1$-score & {R$_{10}$@1} & {R$_{10}$@2} & {R$_{10}$@5} \\
					\hline
					Tri-Attention$_\text{TAdd}$	&\textbf{87.49} ($\pm 0.47$)  & \textbf{87.95} ($\pm 0.27$) &\textbf{90.5} ($\pm 0.3$) & \textbf{95.8} ($\pm 0.06$) & \textbf{99.2} ($\pm 0.05$)	\\
					C-BiAttention$_\text{Add}$	& 86.57 ($\pm 0.15$)  & 87.23 ($\pm 0.07$) 	&90.1 ($\pm 0.4$) & 95.7 ($\pm 0.3$) & \textbf{99.2} ($\pm 0.06$)	\\
					\hline
					Tri-Attention$_\text{TDP}$	&\textbf{87.25} ($\pm 0.11$)  &\textbf{87.83} ($\pm 0.06$) &\textbf{90.3} ($\pm 0.1$) & \textbf{95.9} ($\pm 0.3$) & \textbf{99.3} ($\pm 0.06$)	\\
					C-BiAttention$_\text{DP}$	& 85.94 ($\pm 0.21$)  & 86.81 ($\pm 0.12$) 	&90.1 ($\pm 0.3$) & 95.6 ($\pm 0.2$) & 99.2 ($\pm 0.06$)	\\
					\hline
					Tri-Attention$_\text{TSDP}$	&\textbf{87.23} ($\pm 0.58$)  &\textbf{87.80} ($\pm 0.29$) &\textbf{90.6} ($\pm 0.3$) & \textbf{95.7} ($\pm 0.05$) & 99.2 ($\pm 0.04$)	\\
					C-BiAttention$_\text{SDP}$	& 86.64 ($\pm 0.11$)  & 87.32 ($\pm 0.06$)	&90.2 ($\pm 0.1$) & 95.7 ($\pm 0.2$) & \textbf{99.3} ($\pm 0.04$)	\\
					\hline
					Tri-Attention$_\text{Trili}$	&\textbf{86.72} ($\pm 0.05$)  &\textbf{87.38} ($\pm 0.02$) &90.1 ($\pm 0.08$) & \textbf{95.7} ($\pm 0.09$) & \textbf{99.3} ($\pm 0.04$)	\\
					C-BiAttention$_\text{Bili}$	& 85.94 ($\pm 0.10$)  & 86.82 ($\pm 0.06$)	&\textbf{90.2} ($\pm 0.4$) & 95.7 ($\pm 0.2$) & 99.2 ($\pm 0.05$)	\\
					\hline\hline
				\end{tabular}
				\caption{\label{tabs:Necessity Ablation} Results of contextual query-key interactions vs query-key-context interactions on LCQMC and Ubuntu Corpus V1 Corpus. Means and standard deviations are averaged over five runs.}
			\end{center}
		\end{table*}

		\subsection{Bi-Attention vs Tri-Attention Comparison}
		\label{subsec:ablation}
		Here, we further evaluate the effectiveness of Tri-Attention. Two sets of comparisons are undertaken. The first compares Bi-Attention with Tri-Attention to show the effectiveness of Tri-Attention. The second compares Tri-Attention with a commonly used contextual attention enhancement method in the literature to show the need for capturing explicit query-key-context interactions. As the RACE data costs too much time on running a full set of experiments, we here report the results  on LCQMC and the Ubuntu Corpus V1 corpus. 
		
		\subsubsection{Effectiveness: Bi-Attention vs Tri-Attention Mechanisms}
		
		We evaluate the different effect of the Bi-Attention without contextual information versus the Tri-Attention with context in the same network structure. The experimental results are shown in Table \ref{tabs:Effectiveness Ablation}. Bi-Attention mechanism does not involve contextual information in measuring the interactions. Its other parts are the same as Tri-Attention. We report the results of these two attention mechanisms under four relevance scoring operations: TAdd, TDP, TSDP, and Trili, respectively.  

		Overall, the results show that the Tri-Attention variants consistently outperform the counterparts without context. This corroborates the contribution of Tri-Attention with context, even though the performance varies over different relevance scoring functions.  

		Specifically, on LCQMC, the difference between Bi-Attention and Tri-Attention mechanisms shows the greatest under trilinear operation. Tri-Attention gains extra 1.03\% accuracy and 0.76\% F$_1$-score over Bi-Attention, respectively. For the Ubuntu Corpus V1 corpus, the greatest difference  is associated with SDP, where Tri-Attention makes 0.8\%, 0.2\%, and 0.1\% improvement in terms of evaluation metrics R$_{10}$@1, R$_{10}$@2, and R$_{10}$@5, respectively.
		
		\subsubsection{Necessity: Contextual Query-Key Interactions vs Query-Key-Context Interactions}
		As discussed in Section \ref{sec:relatedwork}, some existing methods also involve contextual information, typically by simple addition or concatenation to target representations. Here, we evaluate the difference of this way from our approaches in Tri-Attention which involves context equivalently to other query and key entities in the attention learning. 
		
		We generate a Bi-Attention variant to realize a contextual attention enhancement method commonly used in the existing work. The contextual information is added (with addition) to each sequence separately to obtain a contextual sequence representation. Then, the context-enhanced sequence representations interact using the standard query-key Bi-Attention mechanism, whose other network settings are consistent with Tri-Attention. This forms the commonly used contextual attention in the literature, abbreviated \textit{C-BiAttention}. 
		
		The experimental results are shown in Table \ref{tabs:Necessity Ablation}. It shows that the performance of C-BiAttention consistently underperforms than Tri-Attention with all four relevance calculators. C-BiAttention with the dot-product-based relevance calculator achieves the lowest performance. On LCQMC, C-BiAttention reduces 1.52\% accuracy and 1.17\% F$_1$-score over Tri-Attention, respectively. Similarly, on the Ubuntu Corpus V1 corpus, the highest performance degradation is with the Additive-based approach. In comparison with Tri-Attention, C-BiAttention decreases 0.4\% R$_{10}$@1, 0.1\% R$_{10}$@2, respectively.
		This experiment verifies that the explicit query-key-context interactions in Tri-Attention outperforms the simple contextual attention by adding or concatenating contextual information to underlying sequences. It also confirms the necessity of learning query-key-context interactions in Tri-Attention.
		
		\begin{figure*}[h]
			\centering
			\subfigure[Tri-Attention$_{\text{TAdd}}$]{
				\begin{minipage}[t]{0.25\linewidth}
					\centering
					\includegraphics[width=1.7in]{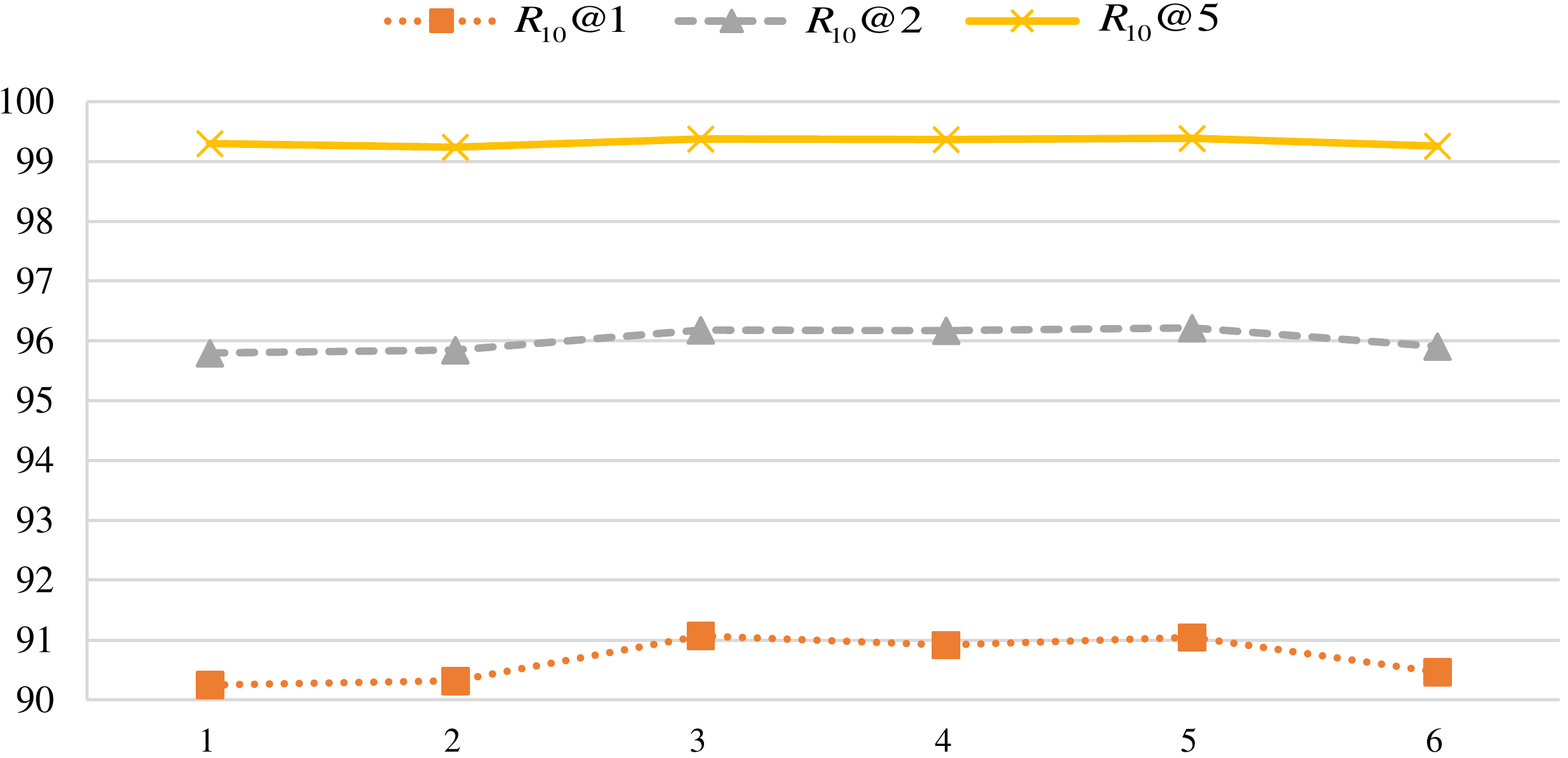}
					\label{fig:add_ubuntu}
				\end{minipage}%
			}%
			\subfigure[Tri-Attention$_{\text{TDP}}$]{
				\begin{minipage}[t]{0.25\linewidth}
					\centering
					\includegraphics[width=1.7in]{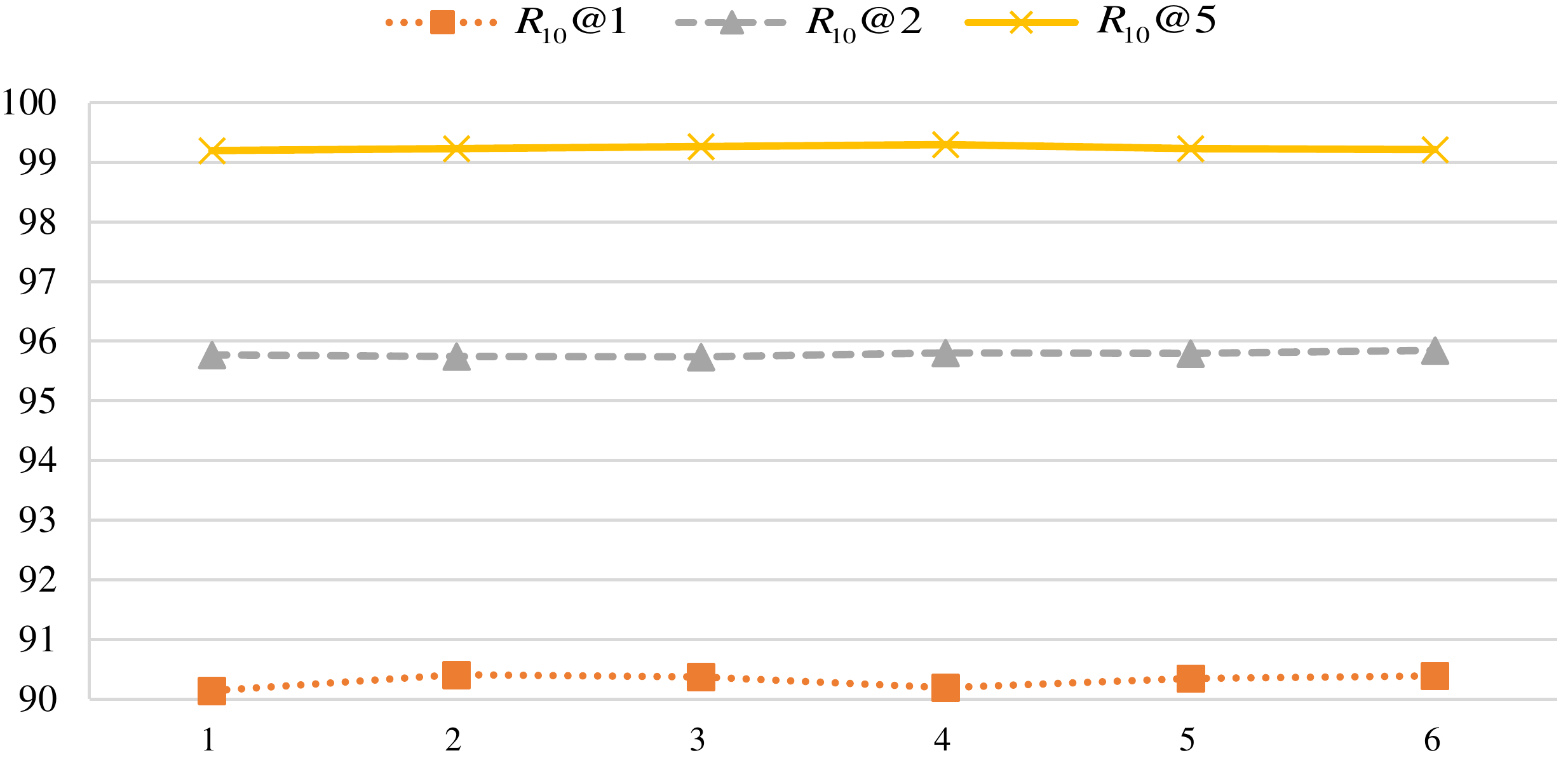}
					\label{fig:dot_ubuntu}
				\end{minipage}%
			}%
			\subfigure[Tri-Attention$_{\text{TSDP}}$]{
				\begin{minipage}[t]{0.25\linewidth}
					\centering
					\includegraphics[width=1.7in]{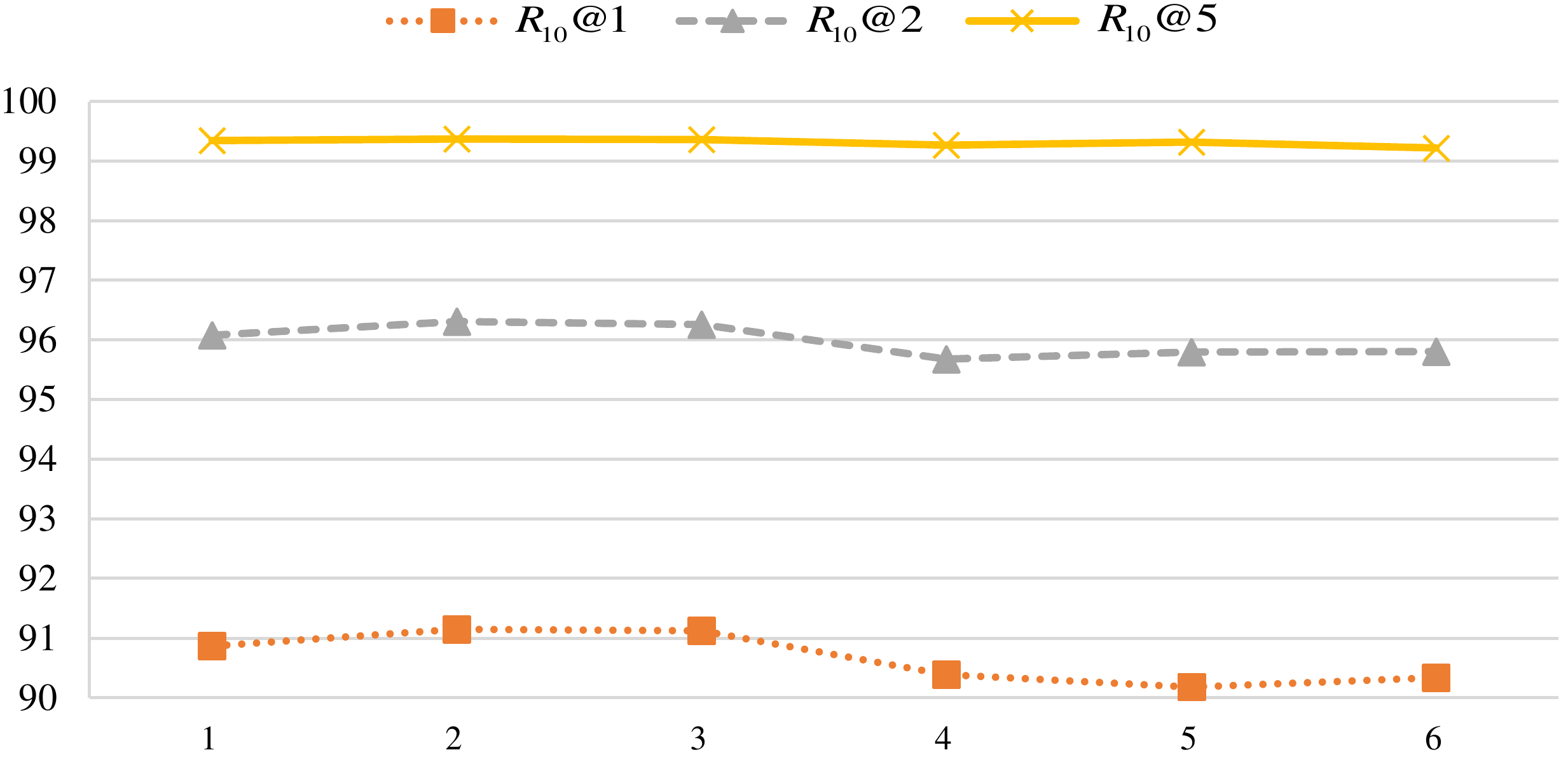}
					\label{fig:sdot_ubuntu}
				\end{minipage}%
			}%
			\subfigure[Tri-Attention$_{\text{Trili}}$]{
				\begin{minipage}[t]{0.25\linewidth}
					\centering
					\includegraphics[width=1.7in]{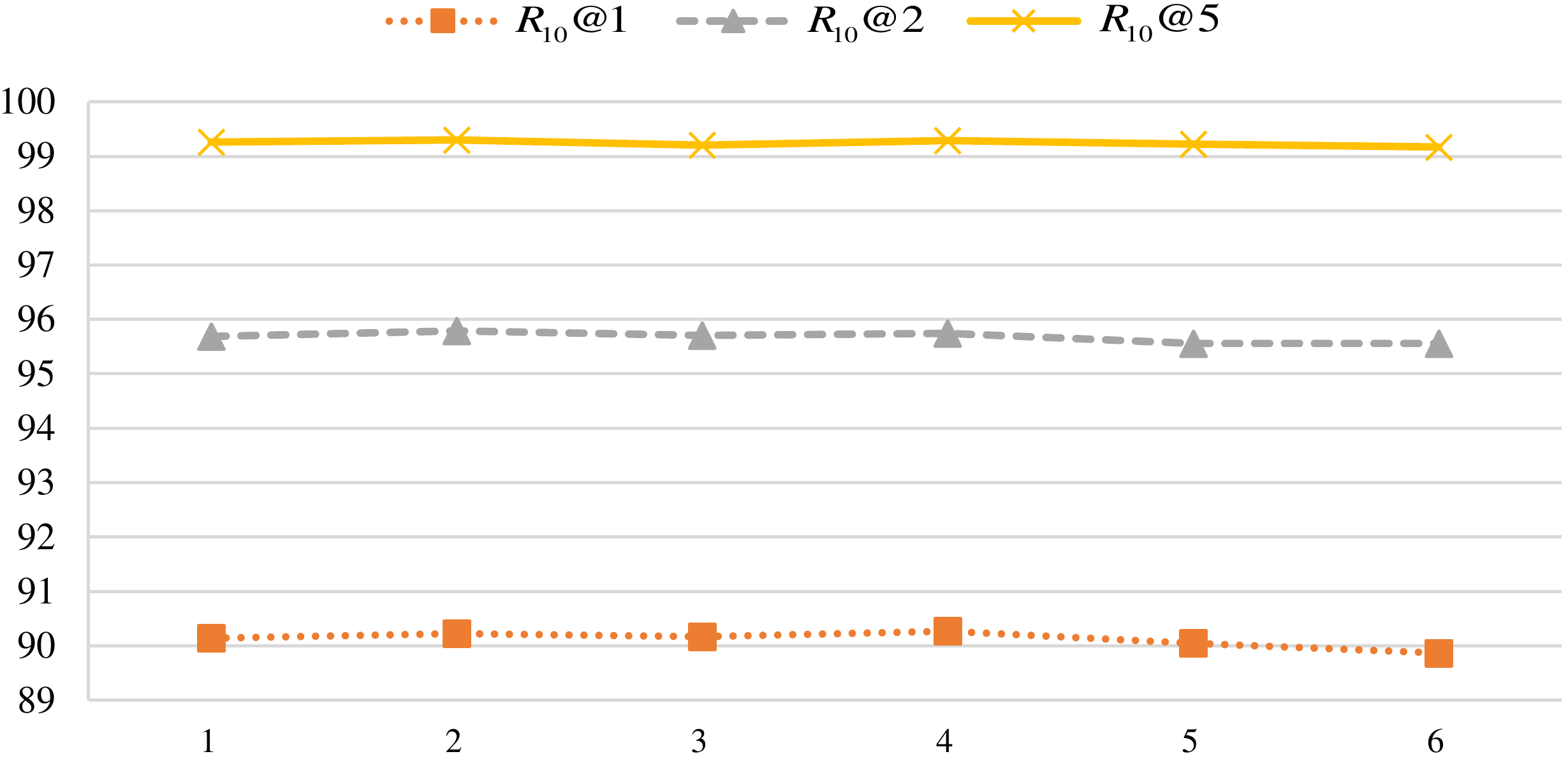}
					\label{fig:gene_ubuntu}
				\end{minipage}%
			}%
			\centering
			\caption{Performance comparison on the Ubuntu Corpus V1 corpus with different number of Tri-Attention layers.}
			\label{fig:6}
		\end{figure*}
		
		\begin{figure*}[h]
			\centering
			\subfigure[Tri-Attention$_{\text{TAdd}}$]{
				\begin{minipage}[t]{0.25\linewidth}
					\centering
					\includegraphics[width=1.7in]{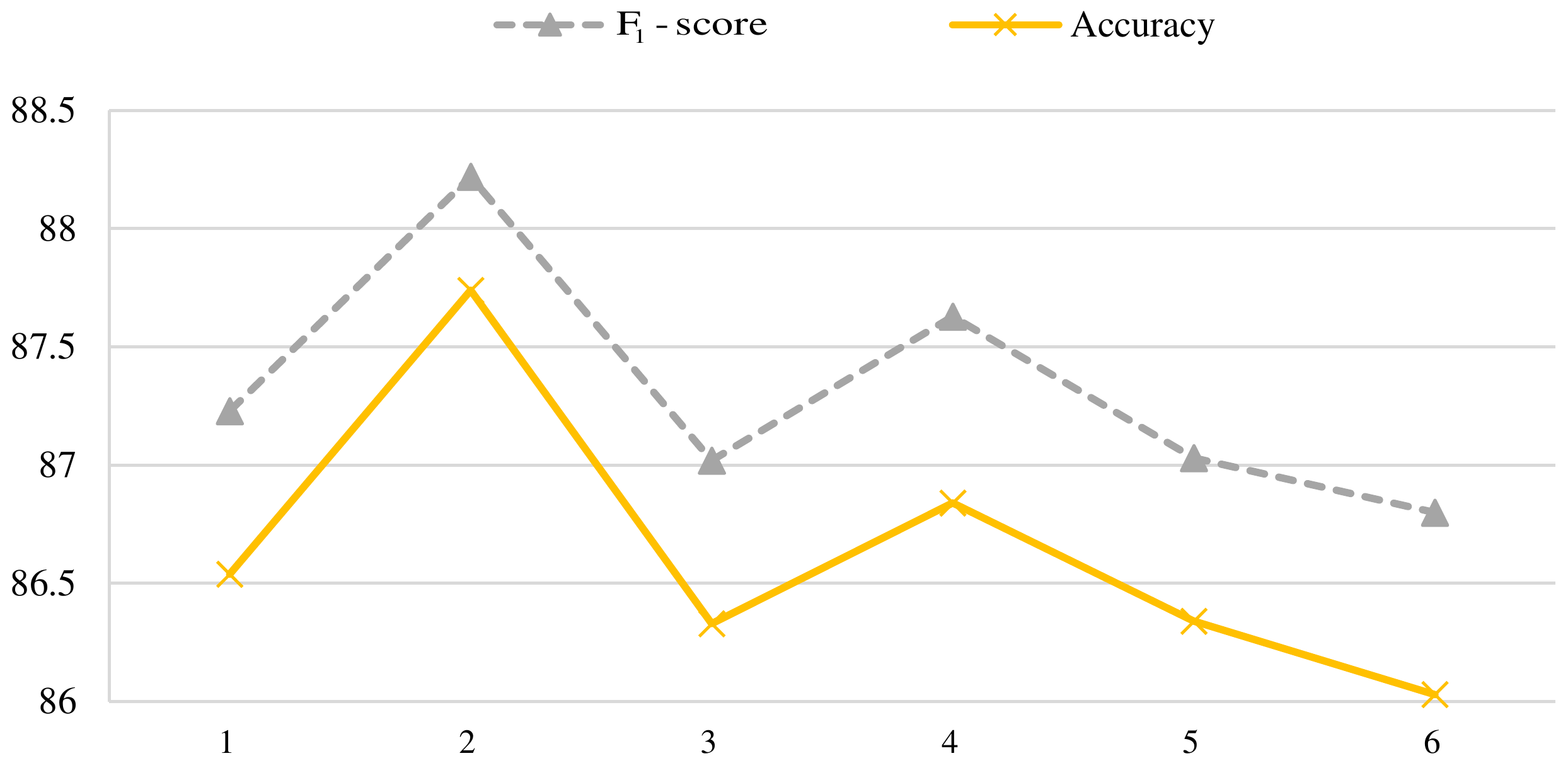}
					\label{fig:add}
				\end{minipage}%
			}%
			\subfigure[Tri-Attention$_{\text{TDP}}$]{
				\begin{minipage}[t]{0.25\linewidth}
					\centering
					\includegraphics[width=1.7in]{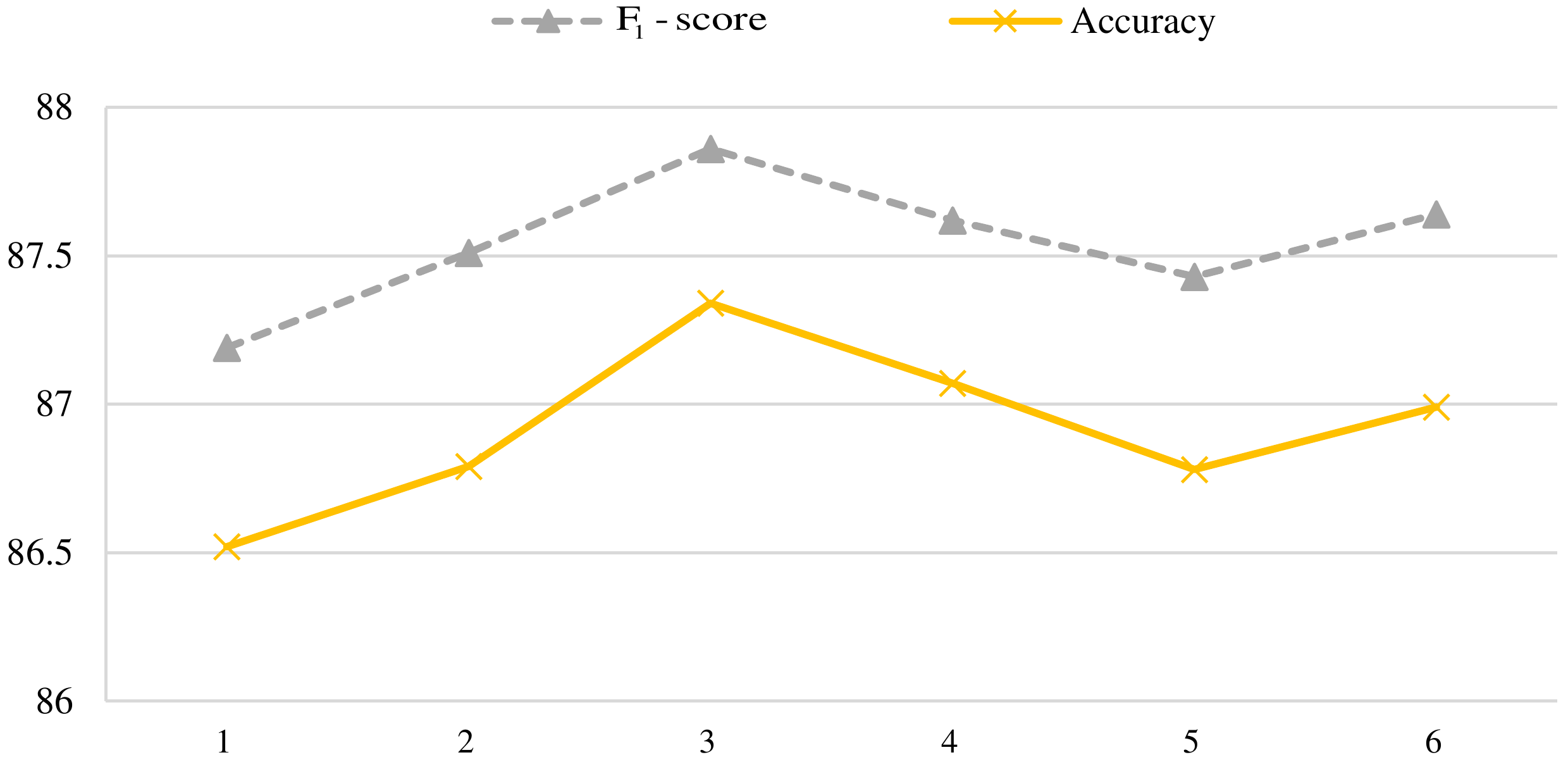}
					\label{fig:dot}
				\end{minipage}%
			}%
			\subfigure[Tri-Attention$_{\text{TSDP}}$]{
				\begin{minipage}[t]{0.25\linewidth}
					\centering
					\includegraphics[width=1.7in]{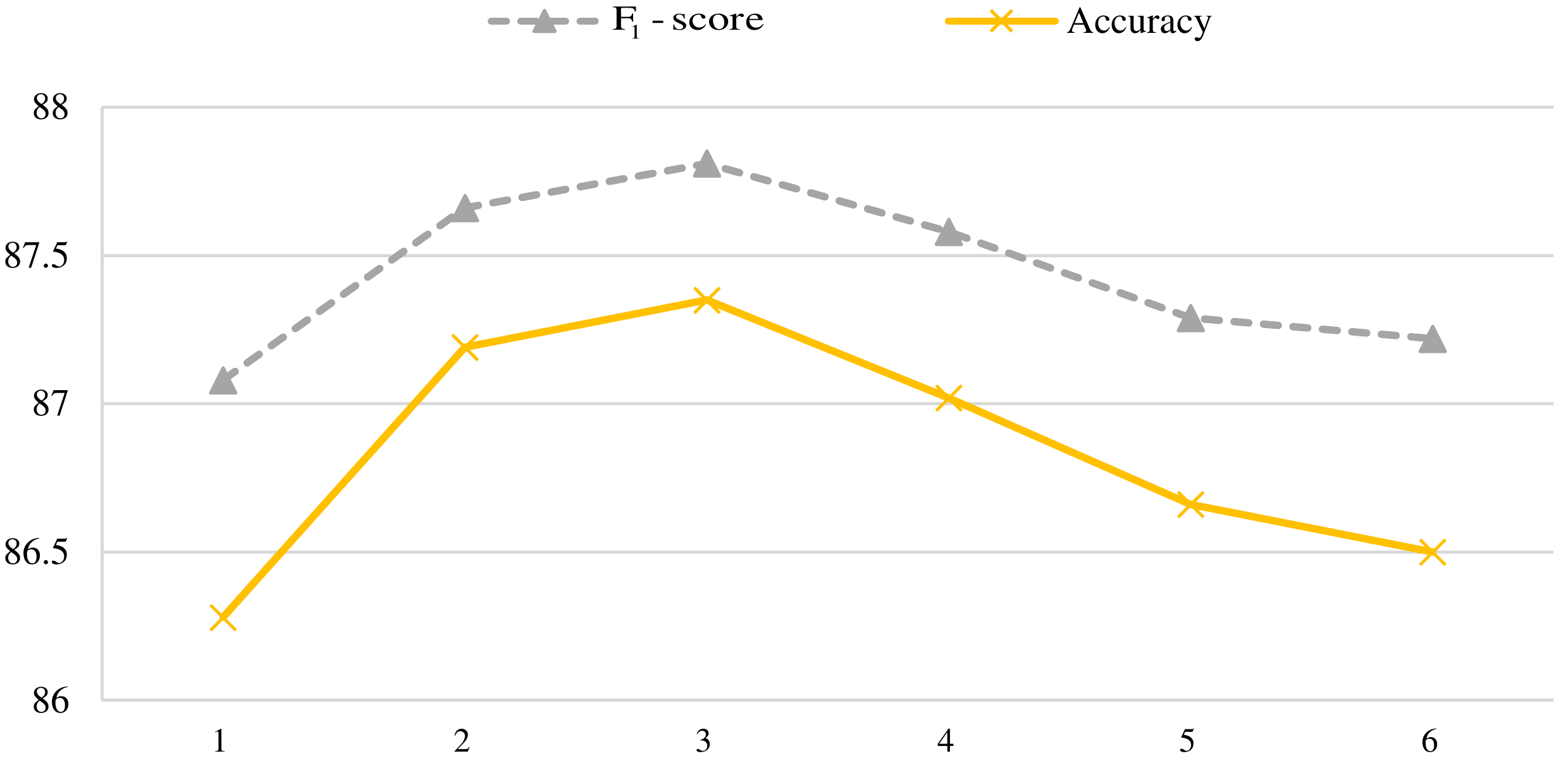}
					\label{fig:sdot}
				\end{minipage}%
			}%
			\subfigure[Tri-Attention$_{\text{Trili}}$]{
				\begin{minipage}[t]{0.25\linewidth}
					\centering
					\includegraphics[width=1.7in]{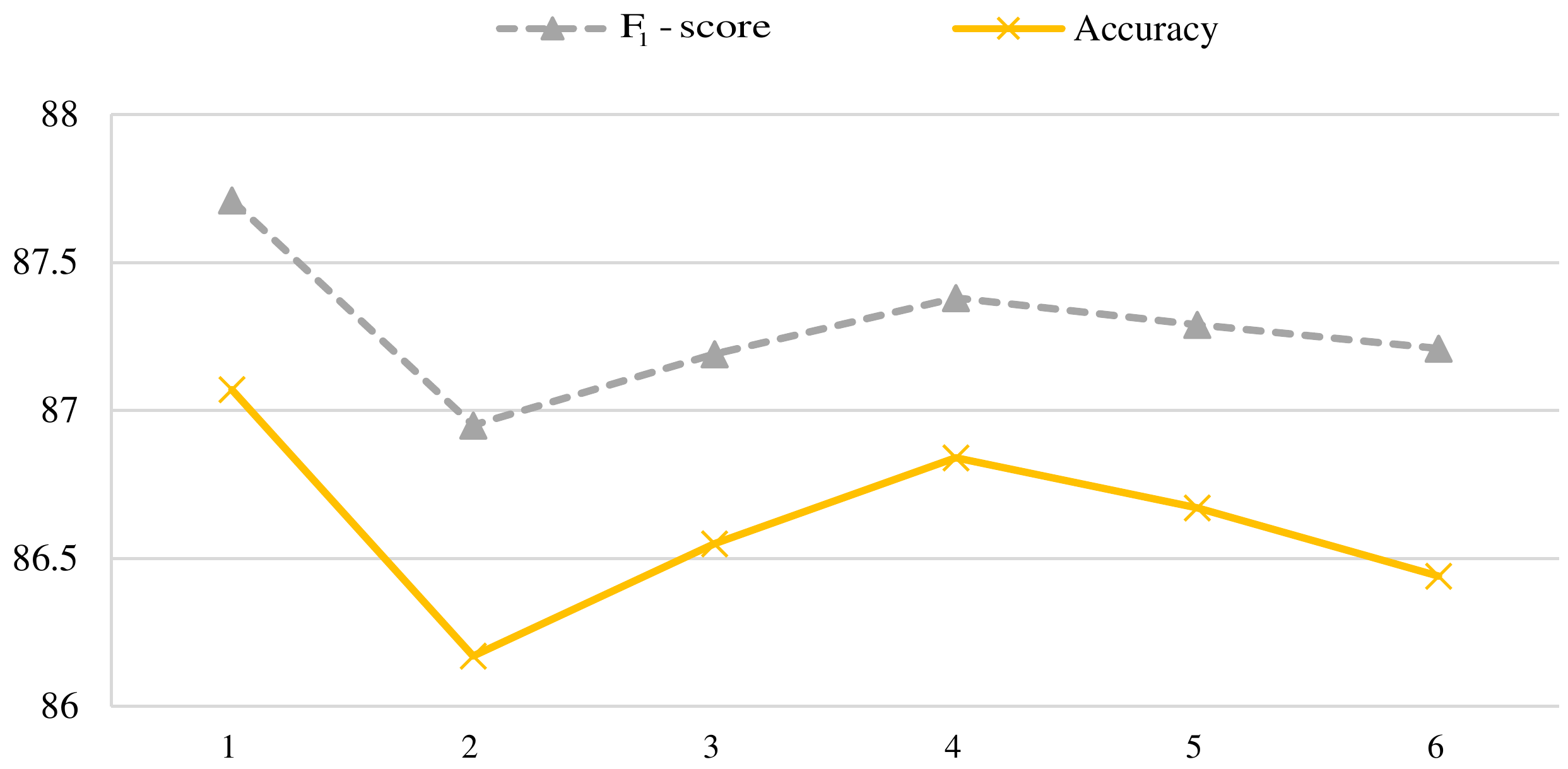}
					\label{fig:gene}
				\end{minipage}%
			}%
			\centering
			\caption{Performance comparison on LCQMC with different number of Tri-Attention layers.}
			\label{fig:5}
		\end{figure*}
		
		\begin{figure*}[h]
			\centering
			\subfigure[Tri-Attention$_{\text{TAdd}}$]{
				\begin{minipage}[t]{0.25\linewidth}
					\centering
					\includegraphics[width=1.7in]{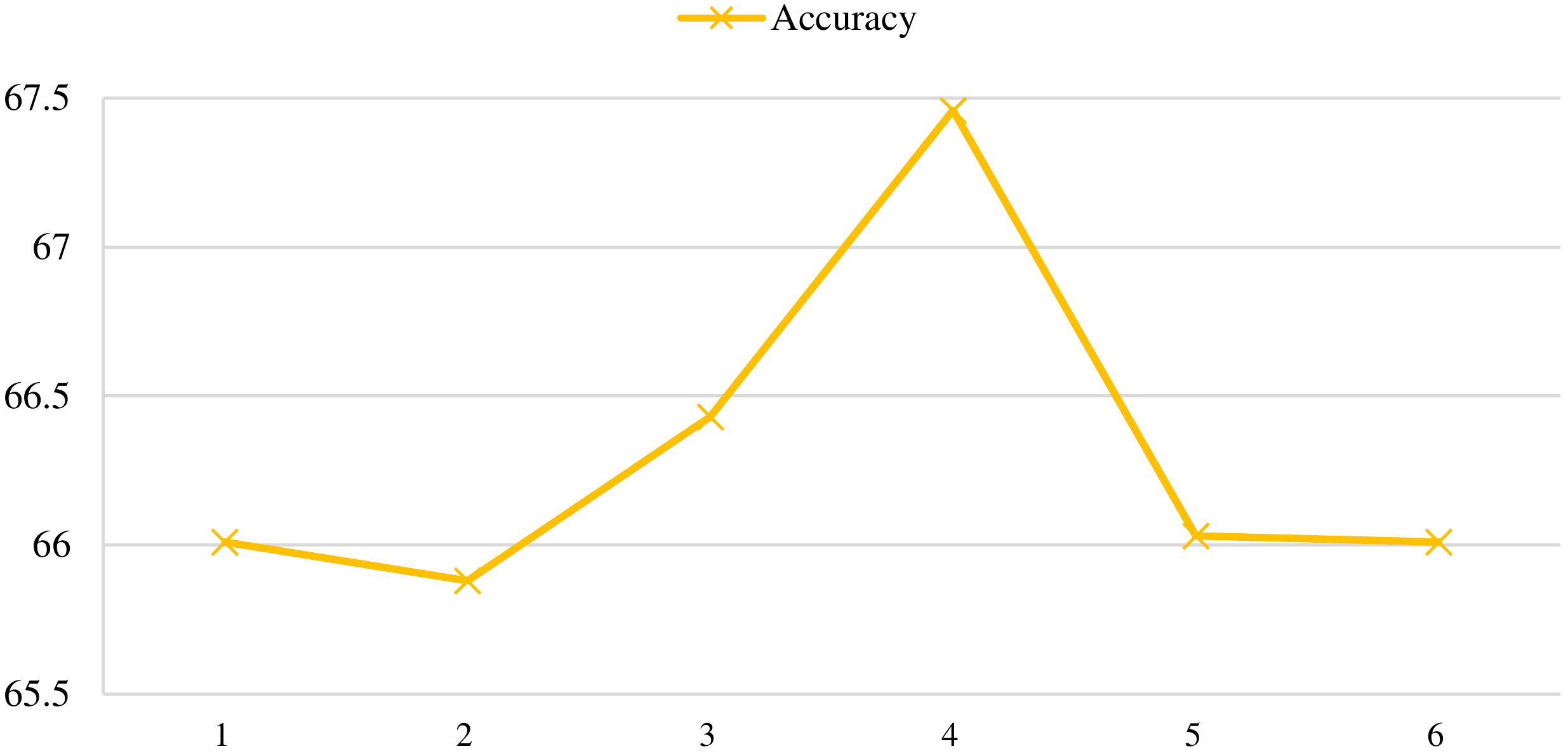}
					\label{fig:add_race}
				\end{minipage}%
			}%
			\subfigure[Tri-Attention$_{\text{TDP}}$]{
				\begin{minipage}[t]{0.25\linewidth}
					\centering
					\includegraphics[width=1.7in]{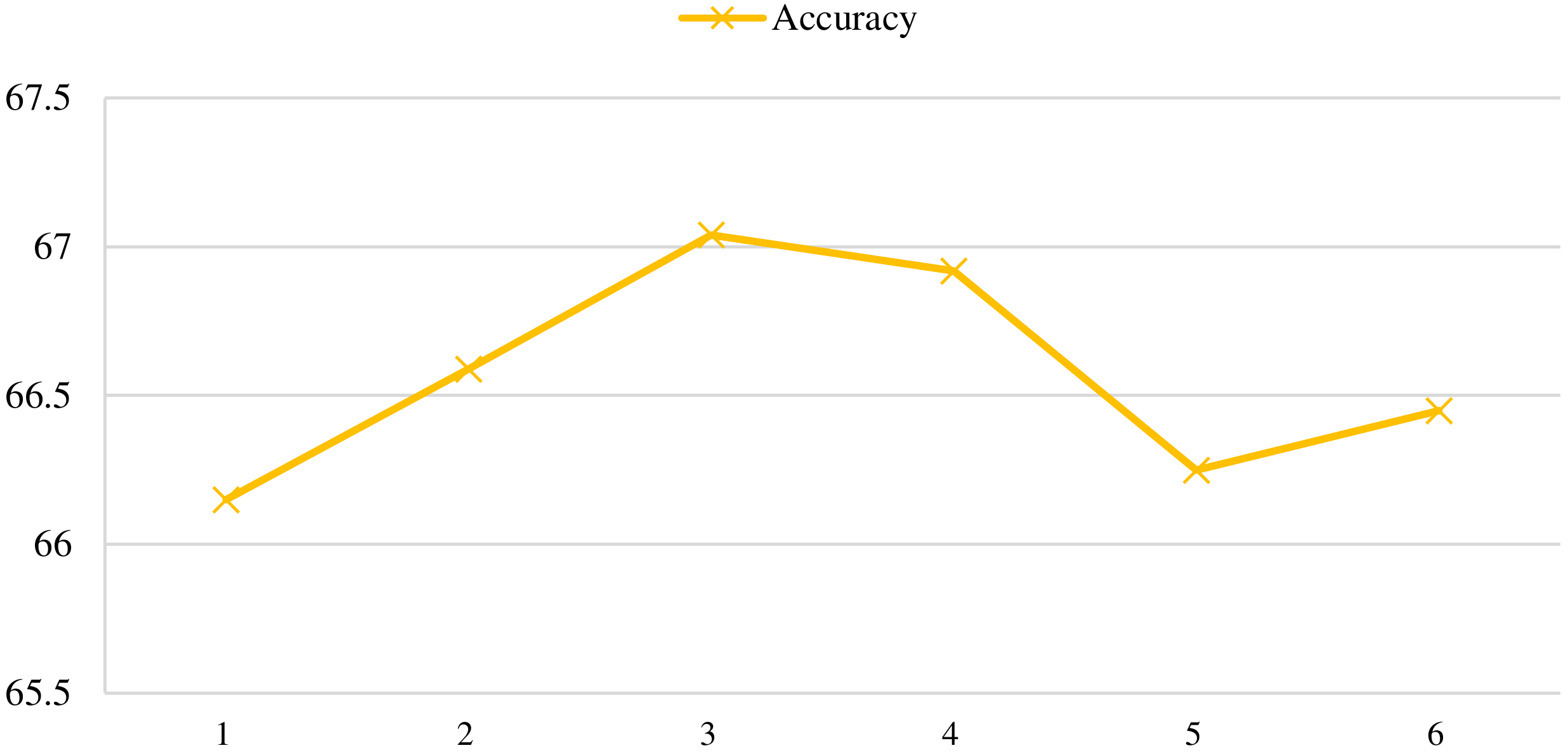}
					\label{fig:dot_race}
				\end{minipage}%
			}%
			\subfigure[Tri-Attention$_{\text{TSDP}}$]{
				\begin{minipage}[t]{0.25\linewidth}
					\centering
					\includegraphics[width=1.7in]{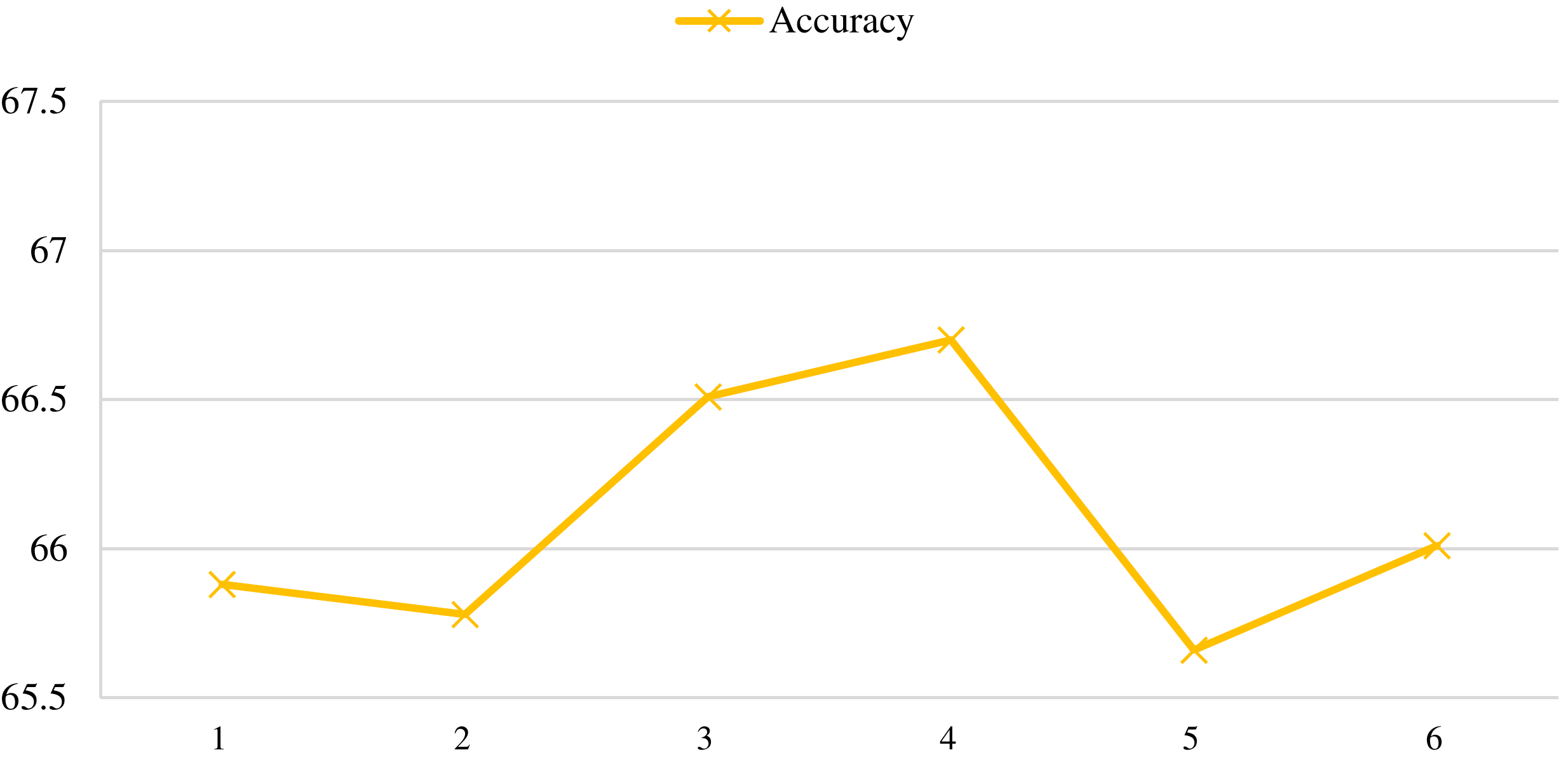}
					\label{fig:sdot_race}
				\end{minipage}%
			}%
			\subfigure[Tri-Attention$_{\text{Trili}}$]{
				\begin{minipage}[t]{0.25\linewidth}
					\centering
					\includegraphics[width=1.7in]{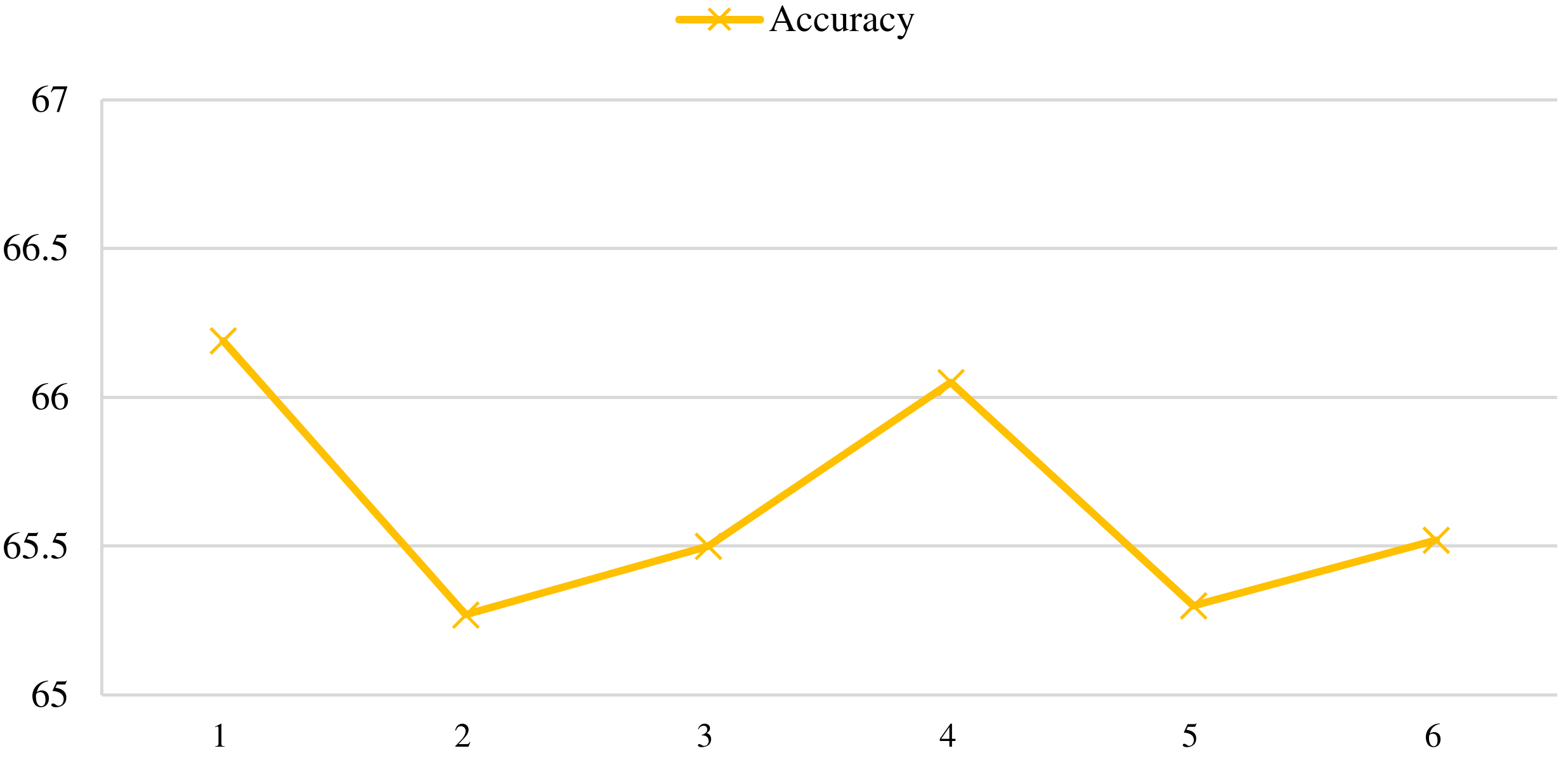}
					\label{fig:gene_race}
				\end{minipage}%
			}%
			\centering
			\caption{Performance comparison on RACE with different number of Tri-Attention layers.}
			\label{fig:7}
		\end{figure*}

		\subsection{Hyperparameter Analysis: Tri-Attention Layers}
		TAN in Fig. \ref{T-A} is a stackable structure, where the number of Tri-Attention layers can be dynamically adjusted according to learning tasks. In addition, In addition, different relevance score calculation methods in Eqs. \eqref{formula:concat score context}-\eqref{formula:general score context} also affect  the number of Tri-Attention layers for a learning task. We thus test the effect of the number of Tri-Attention layers. 
		
		Fig. \ref{fig:6} - \ref{fig:7} show the experimental results of 
        retrieval-based dialogue, sentence semantic matching, and machine reading comprehension tasks on Ubuntu Corpus V1, LCQMC, and RACE, respectively. The best performance of Tri-Attention corresponds to different relevance score calculation methods for different tasks. This indicates that, when applying Tri-Attention to a special task, all four relevance calculation methods should be tried before achieving the best result.    
		In addition, the number of Tri-Attention layers depends on the relevance calculation methods and learning tasks. Thus, when applied to a specific task, the best number of Tri-Attention layers needs to be tuned for each relevance calculation method.  

		\subsection{TAN Case Study}
		Lastly, we illustrate the prediction results of TAN with Tri-Attention on the LCQMC data for sentence semantic matching.
		We only illustrate the additive-based relevance calculation for Tri-Attention as the previous experiments show its better stability. We reapply the three variants used in Section \ref{subsec:ablation}: Tri-Attention$_{\text{Add}}$, Bi-Attention$_{\text{Add}}$ and C-BiAttention$_{\text{Add}}$, respectively.
		To verify the advantages of Tri-Attention, the prediction results of all three models were statistically analyzed. 
		
		Specifically, we conduct a statistical analysis when only Tri-Attention makes correct prediction while the other two models predict incorrectly.
		In the test set of LCQMC, there are 151 pieces of data conforming to the above situation, among which 49 pieces are positive and 102 pieces are negative.
		This suggests that Tri-Attention is better at identifying negative data.
		The instance in Table \ref{tabs:case} illustrates the result. 
		
		\begin{table}[ht]\small
			\begin{center}
				\begin{tabular}{c|c|c}
					\hline\hline
					& Data &  Label \\
					\hline
					\multirow{2}*{S$_1$}& 哪些浏览器可以\textcolor{red}{看}电影 	&\multirow{4}*{0}\\
					& (En: Which browsers can \textcolor{blue}{play} movies)	&\\
					\cline{1-2}
					\multirow{2}*{S$_2$}& 什么浏览器可以\textcolor{red}{下}电影	&\\
					& (En: Which browsers can \textcolor{blue}{download} movies)	&\\
					\hline\hline
				\end{tabular}
			\end{center}
			\caption{\label{tabs:case} Case study examples from LCQMC.}
		\end{table}
		
		For both S$_1$ and S$_2$, the gold label is 0, which means that their meanings are different. Our Tri-attention can make a correct judgement, while the others cannot. Since there are many overlap words in S$_1$ and S$_2$, if a model judges the relations between these two sentences only on the basis of word-level relevance without the contextual information, it is easy to be misled and draw a wrong conclusion. 
		This probably explains why the Bi-Attention$_{\text{Add}}$ method fails to correctly predict the relation between S$_1$ and S$_2$.
		As for C-BiAttention$_{\text{Add}}$, although it also involves context, its contextual information does not participate in the interaction with queries and keys, thus making no direct impact on the relevance score between sentences.
		While Tri-Attention directly involves context in the interactions between sentences, contributing to better prediction.

		\section{Conclusions}
        \label{sec:conclusion}
		Contextual information has shown essential in many learning tasks. The great success of attention mechanisms does not necessarily involves contextual information. In neural NLP, increasing work on contextual attention learning typically concatenates contextual features into underlying targets such as sequences, then the standard query-key-based Bi-Attention mechanisms calculate relevance scores on the tokenized contextual sequence representations. In this paper, a novel query-key-context-interactive Tri-Attention mechanism explicitly captures the interactions between query, key and context. We derive four query-key-context relevance calculation methods for Tri-Attention using tensor algebraic techniques. Intensive experiments on different NLP tasks show that Tri-Attention-based networks can serve as a general attention framework, which outperforms most state-of-the-art non-attention, standard Bi-Attention, contextual Bi-Attention, and pretrained language models with attention. Our future work includes evaluating Tri-Attention in other NLP tasks and Tri-Attention without pretrained BERT, and exploring more effective tensor algebraic implementations for 
		for interactions with $n>3$ factors.


		
		%

		


		\ifCLASSOPTIONcaptionsoff
		\newpage
		\fi

		\bibliographystyle{IEEEtran}
		\bibliography{mybibfile}
		
		
		\begin{IEEEbiography}[{\includegraphics[width=1in,height=1.25in,clip,keepaspectratio]{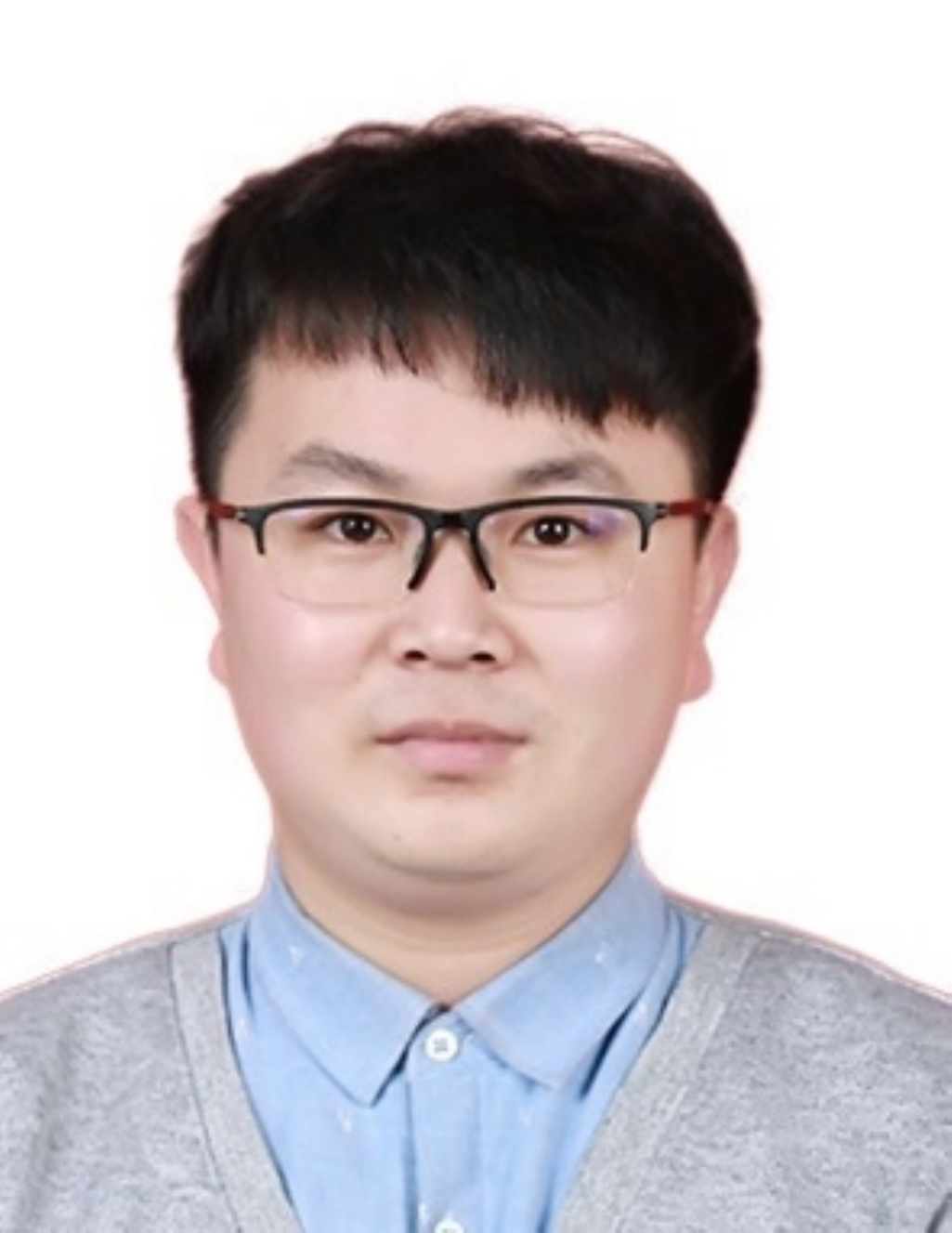}}]{Rui Yu}
        received the master’s degree in computer application technology from the Qilu University of Technology (Shandong Academy of Sciences), Jinan, China. He is pursuing the Ph.D. degree at Huazhong University of Science and Technology, Wuhan, China.       His research interests include text semantic matching and question answering system. He may be contacted at rui.yu1996@foxmail.com.
        \end{IEEEbiography}
        
		\begin{IEEEbiography}[{\includegraphics[width=1in,height=1.25in,clip,keepaspectratio]{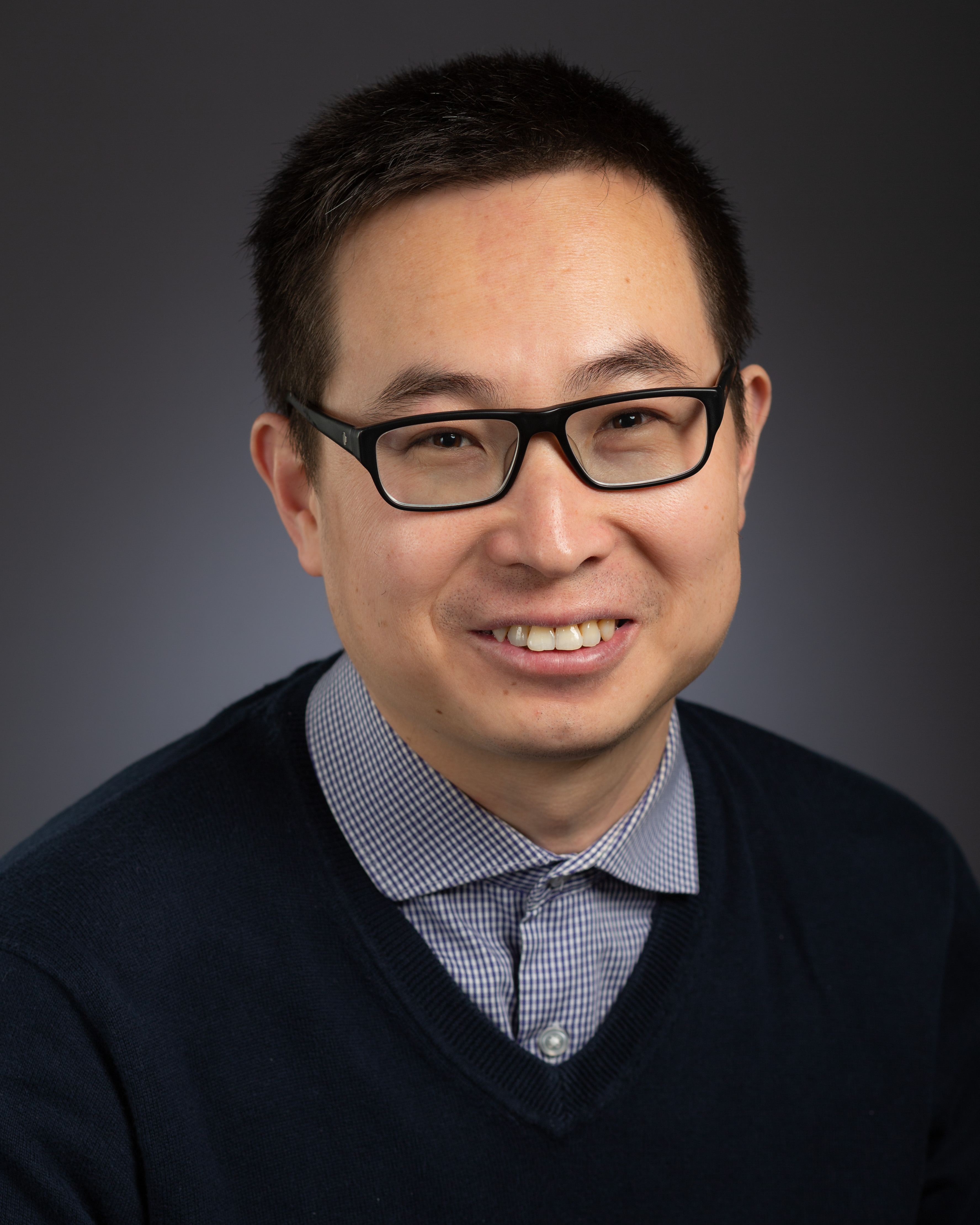}}]{Yifeng Li}
		received the Ph.D. in Computer Science from the University of Windsor, Canada. 
		He is an assistant professor and Canada Research Chair (Tier 2) in Machine Learning for Biomedical Data Science at the Department of Computer Science, Department of Biological Sciences, and Centre for Biotechnology, Brock University. 
		His research interests include neural networks, machine learning, data science, optimization, bioinformatics, and chemoinformatics. He may be contacted at yli2@brocku.ca.

        \end{IEEEbiography}
        
        \begin{IEEEbiography}[{\includegraphics[width=1in,height=1.25in,clip,keepaspectratio]{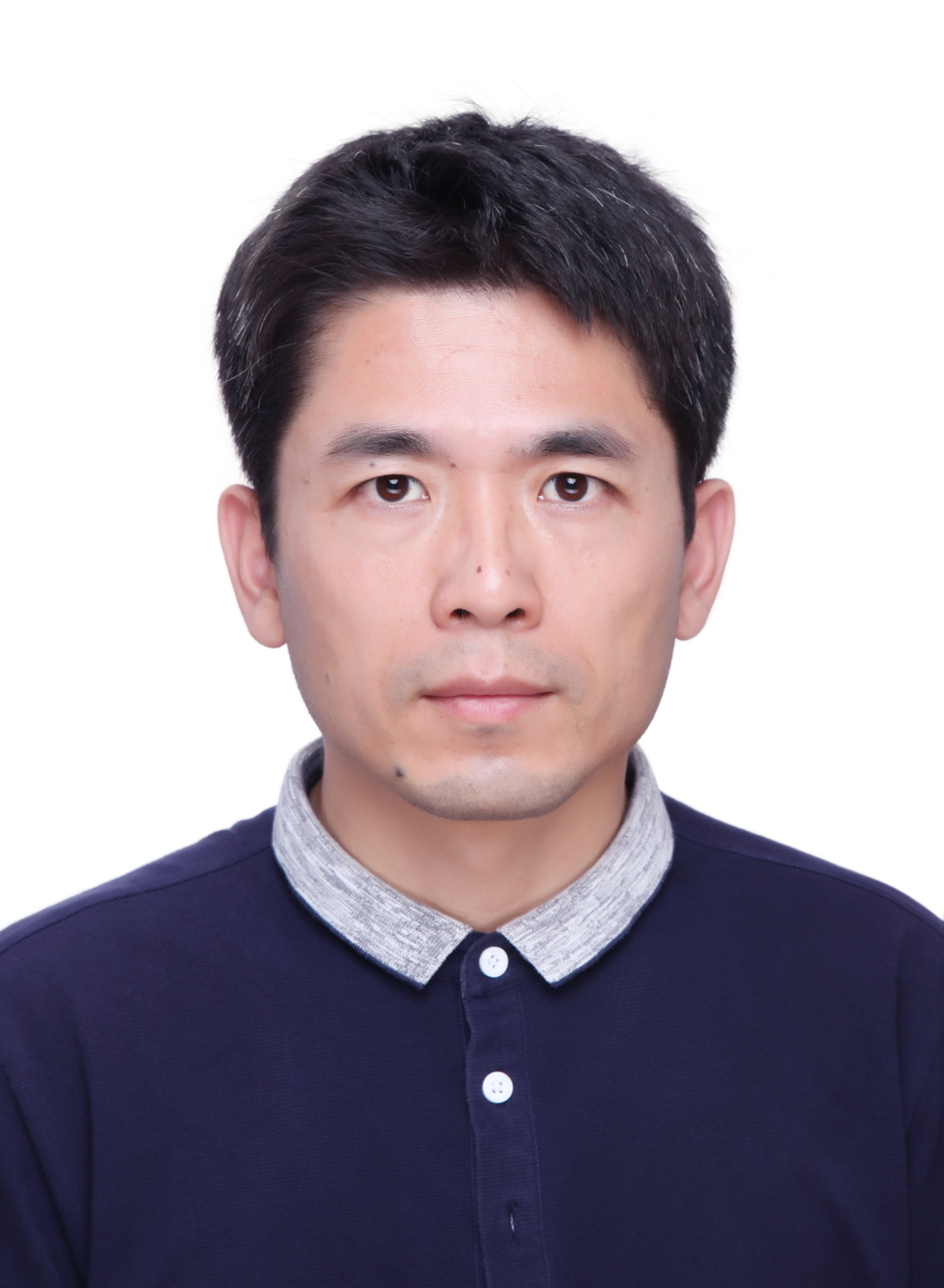}}]{Wenpeng Lu} received the Ph.D. degree in computer applications from the Beijing Institute of Technology, Beijing, China.
        He is a professor with the Department of Computer Science and Technology, Qilu University of Technology (Shandong Academy of Sciences), Jinan, China. 
        His research interests include natural language processing, machine learning, and their enterprise applications. He is a member of IEEE, CCF and CIPS. He may be contacted at wenpeng.lu@qlu.edu.cu.
        \end{IEEEbiography}
		
        \begin{IEEEbiography}[{\includegraphics[width=1in,height=1.25in,clip,keepaspectratio]{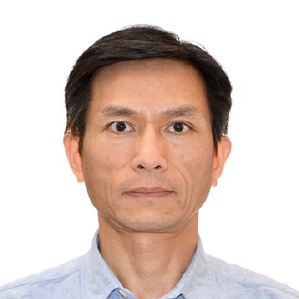}}]{Longbing Cao} received a Ph.D. degree in pattern recognition and intelligent systems and another Ph.D. in computing sciences. He is a professor and an ARC Future Fellow (level 3) at the University of Technology Sydney. His research interests include AI, data science, machine learning, behavior informatics, and enterprise innovation. He is the EICs of IEEE Intelligent Systems and Springer's JDSA. He may be contacted at longbing.cao@uts.edu.au.
        \end{IEEEbiography}
		
	\end{CJK*}
\end{document}